%% file: acl_latex.tex
\documentclass[11pt]{article}

\usepackage[final]{acl}

\usepackage{times}
\usepackage{latexsym}
 
\usepackage[T1]{fontenc}

\usepackage[utf8]{inputenc}

\usepackage{microtype}

\usepackage{inconsolata}

\usepackage{graphicx}

\usepackage{amsfonts}
\usepackage{amsmath}
\usepackage{booktabs}
\usepackage{multirow}
\usepackage{makecell}
\usepackage{xspace}
\usepackage{xcolor}
\usepackage{cleveref}
\title{Language-Grounded Multi-Domain Image Translation via Semantic Difference Guidance}

\author{
 \textbf{Jongwon Ryu\textsuperscript{1}\thanks{Equal contribution}},
 \textbf{Joonhyung Park\textsuperscript{2}$^\ast$},
 \textbf{Jaeho Han\textsuperscript{1}},
 \textbf{Yeong-Seok Kim\textsuperscript{2}},
\\
 \textbf{Hye-rin Kim\textsuperscript{2}},
 \textbf{Sunjae Yoon\textsuperscript{3}},
 \textbf{Junyeong Kim\textsuperscript{1}}
\\
\\
 \textsuperscript{1}Department of Artificial Intelligence, Chung-Ang University,\\
 \textsuperscript{2}Hyundai Mobis,
 \textsuperscript{3}Korea Advanced Institute of Science and Technology
\\
 \small{
   \textbf{Correspondence:} \href{mailto:email@domain}{Junyeongkim@cau.ac.kr}
 }
}

\begin{document}
\maketitle

\begin{abstract}
Multi-domain image-to-image translation requires grounding semantic differences expressed in natural language prompts into corresponding visual transformations, while preserving unrelated structural and semantic content. Existing methods struggle to maintain structural integrity and provide fine-grained, attribute-specific control, especially when multiple domains are involved. We propose LACE (Language-grounded Attribute-Controllable Translation), built on two components: (1) a GLIP-Adapter that fuses global semantics with local structural features to preserve consistency, and (2) a Multi-Domain Control Guidance mechanism that explicitly grounds the semantic delta between source and target prompts into per-attribute translation vectors, aligning linguistic semantics with domain-level visual changes.
Together, these modules enable compositional multi-domain control with independent strength modulation for each attribute. Experiments on CelebA(Dialog) and BDD100K demonstrate that LACE achieves high visual fidelity, structural preservation, and interpretable domain-specific control, surpassing prior baselines. This positions LACE as a cross-modal content generation framework bridging language semantics and controllable visual translation.
\end{abstract}



\section{Introduction}

Image-to-image (I2I) translation is a fundamental task in computer vision that aims to alter specific visual attributes of an image while preserving its structural and semantic integrity~\cite{huang2018multimodal, lee2018diverse}. It has a wide range of applications, including facial attribute editing~\cite{choi2018stargan}, weather or time simulation in driving scenes~\cite{sun2022shift}, and artistic style transfer~\cite{zhang2023inversion,zhang2024towards,zhang2025u}. In many real-world settings, images are influenced by multiple entangled domain attributes (e.g., "snowy night city"), which calls for models capable of modifying several attributes simultaneously without disrupting non-target content, and crucially, grounding semantic differences expressed in natural language prompts into corresponding visual transformations.

\begin{figure}[t]
    \centering
    \includegraphics[width=1\linewidth]{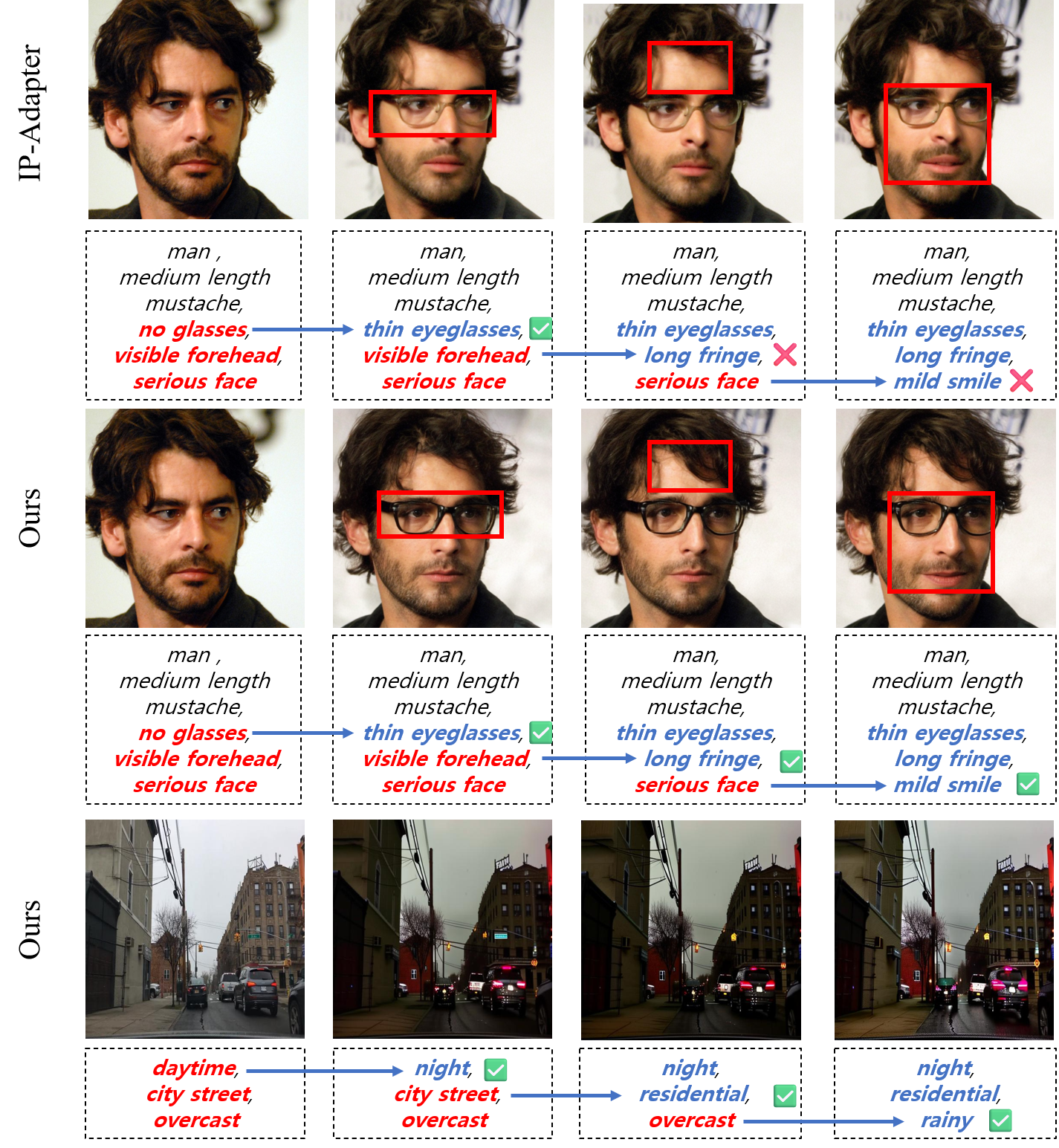}
    \caption{
Multi-domain translation results with progressive attribute modifications.
Our method enables compositional editing of multiple attributes while preserving non-target regions and structural consistency.
    }
    \label{fig:teaser}
\end{figure}

Compared to traditional GAN-based methods~\cite{goodfellow2014generative}, diffusion-based generative models provide more stable training and higher-quality results in image-to-image translation~\cite{saharia2022palette, rombach2022high}.
Nevertheless, current diffusion frameworks still struggle with controllable and structure-preserving multi-attribute translation.
Most are designed for single-attribute editing and rely heavily on text prompts or class labels~\cite{nichol2021glide, kim2022diffusionclip}, which limits their flexibility in multi-domain scenarios.
Some recent approaches attempt to preserve overall image content by inverting inputs into noise and leveraging attention-based mechanisms for domain-specific attribute editing~\cite{mokady2023null, hertz2022prompt}, but they often fail to retain fine-grained structures such as small yet semantically critical elements in complex scenes~\cite{gómez2023UrbanSyn}.
In addition, they lack mechanisms for prompt-driven per-attribute control and fine-grained adjustment of translation strength~\cite{tumanyan2023plug, cao2023masactrl}.
Their reliance on segmentation masks~\cite{choi2020stargan} or spatial annotations~\cite{chen2017deeplabv3, ren2016fasterrcnn} further limits applicability in unstructured real-world settings.
Consequently, existing models often fail to achieve precise and coherent multi-domain translation. As shown in Figure~\ref{fig:teaser}, they yield entangled or distorted outputs when editing multiple attributes simultaneously, underscoring the need for a framework that preserves structure while enabling prompt-level controllability.

To overcome these limitations, we propose LACE (Language-grounded Attribute-Controllable Translation), a diffusion-based framework for multi-domain I2I translation that enables attribute-wise control, including selective editing and per-domain strength modulation. LACE performs multi-attribute translation without relying on region masks or supervision, while preserving the structural integrity of the input image. 
The LACE introduces two key components:
(1) a Global-Local Image Prompt Adapter (GLIP-Adapter) that extracts visual cues by combining global semantics and local structures from a source image, and
(2) a Multi-Domain Control Guidance (MCG) module that explicitly grounds the semantic delta between source and target prompts into noise-space translation vectors, thereby aligning linguistic semantics with domain-level visual changes.
This design allows our model to support targeted editing, compositional prompt control, and domain-specific strength adjustment, making it suitable for complex real-world scenarios involving entangled visual factors.

We validate our approach on CelebA(Dialog)~\cite{jiang2021talkedit}, which involves fine-grained facial attribute editing (e.g., age, gender), and BDD100K~\cite{yu2020bdd100k}, which requires broader domain-level translation across scene factors such as weather and time.
This choice of datasets allows us to demonstrate the versatility of our method across both local attribute manipulation and global background/style translation.
Our experiments involve up to three simultaneous attribute translations per image, and the results show that our method surpasses prior work in translation fidelity, structural consistency, and interpretability of domain-level control, establishing a strong benchmark for controllable diffusion-based multi-domain image translation.

\section{Related work}

\subsection{Conditional Diffusion for I2I Translation}
Diffusion models have recently emerged as a powerful alternative to GANs for image-to-image (I2I) translation, offering improved training stability and high-quality outputs~\cite{ho2020denoising, saharia2022palette}. Conditional diffusion methods such as DiffEdit~\cite{meng2021sdedit}, Pix2PixDiff~\cite{tumanyan2023plug}, and ControlNet~\cite{zhang2023adding} enable task-guided or paired I2I generation by injecting structural inputs like edges or pose maps. However, these approaches are primarily designed for one-to-one or task-specific translation, and often lack scalability to more general multi-domain or compositional scenarios. While models like DiT-Image2Image incorporate CLIP guidance to align with visual semantics, they do not explicitly model domain-aware or attribute-wise transformation, limiting their controllability in multi-attribute editing tasks.


\begin{figure*}[t]
    \centering
    \includegraphics[width=\textwidth]{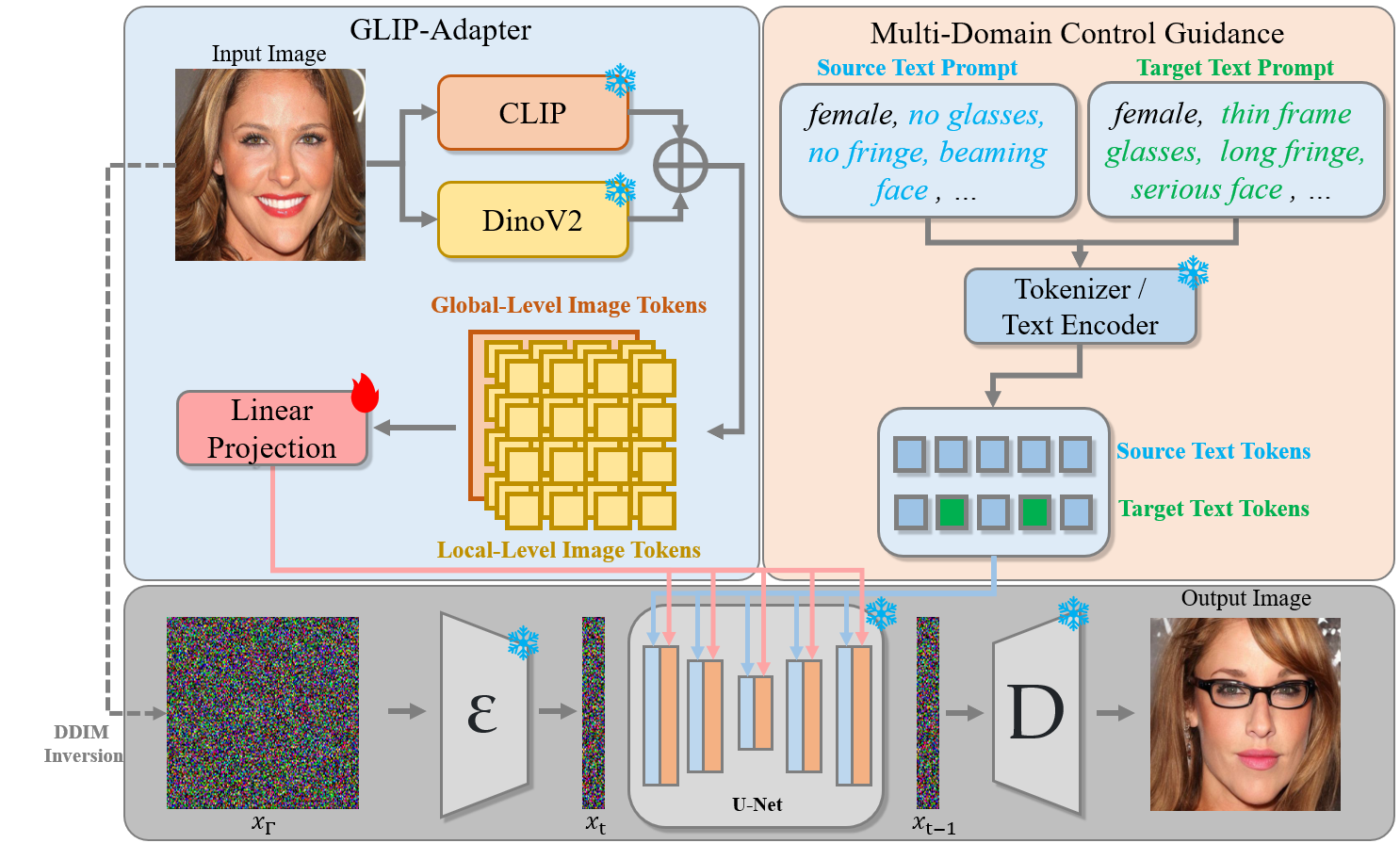}
    \caption{
    Overview of our proposed multi-domain I2I translation framework.
    The GLIP-Adapter projects global (CLIP) and local (DINOv2) features into a linear projection,
    which together with the source text prompt is used to train the U-Net’s cross-attention layers.
    At inference, the MCG module leverages source–target prompt differences to guide controllable multi-attribute translation.
    }

    \label{fig:framework}
\end{figure*}

\subsection{Prompt-Based Guidance and Domain Control}
Classifier-Free Guidance (CFG)~\cite{ho2022classifier} introduced a simple yet effective method for prompt-based conditioning in diffusion models, interpolating between unconditional and conditional predictions. While widely adopted, CFG is inherently limited to single-attribute control and lacks compositionality. Follow-up methods such as Blended CFG~\cite{hertz2022prompt} and FlexiDiffusion~\cite{cao2023masactrl} extend this idea by mixing multiple prompt embeddings, but without explicitly disentangling or controlling individual domain attributes.

Image-conditioned adapters such as IP-Adapter~\cite{ye2023ip} and T2I-Adapter~\cite{mou2024t2i} use CLIP-based embeddings from a reference image to preserve identity or style during generation. However, these approaches offer limited control over which attributes to modify and lack explicit mechanisms for modeling domain-wise transformation strength or direction.

\section{Method}

\subsection{Preliminaries}

Diffusion models~\cite{sohl2015deep, ho2020denoising, rombach2022high} are generative models that learn to reverse a stochastic noising process by predicting the noise added to clean data. Given an image $x_0$, a forward process progressively adds Gaussian noise to obtain a noisy image $x_t$ at time step $t$. The reverse process, parameterized by $\epsilon_\theta$, is trained to predict the added noise $\epsilon$ given the noisy input $x_t$, a conditioning signal $c$, and timestep $t$. The training objective is the denoising score matching loss:

\begin{equation}
    \mathcal{L} = \mathbb{E}_{x_0, \epsilon \sim \mathcal{N}(0, I), c, t} \left\| \epsilon - \epsilon_\theta(x_t, c, t) \right\|^2,
    \label{eq:ddpm_loss}
\end{equation}

where $c$ can be interpreted as a linguistic condition, where semantic differences across prompts define the attribute shifts to be grounded in the visual domain.

In multi-domain image-to-image (I2I) translation, the goal is to selectively modify specific domain attributes (e.g., weather, expression, or lighting) of an input image while preserving its structural layout and semantic content. This task involves three key challenges: (1) isolating and editing only the intended target attributes, (2) maintaining the integrity of non-target regions, and (3) enabling compositional and fine-grained control across multiple domain axes.

Previous diffusion-based translation methods typically rely on single text prompts or global visual features, limiting their applicability in multi-attribute translation. Approaches such as ControlNet~\cite{zhang2023adding} or IP-Adapter~\cite{ye2023ip} introduce strong spatial or identity conditioning but lack explicit mechanisms for prompt-driven, compositional control across domains.
To overcome these limitations, we propose a LACE (Language-grounded Attribute-Controllable Translation). An overview of our proposed architecture is shown in Figure~\ref{fig:framework}.

\subsection{Global-Local Image Prompt Adapter (GLIP-Adapter)}

To preserve structural and semantic details during translation, we introduce the Global-Local Image Prompt Adapter (GLIP-Adapter), which injects visual prompts from a source image into the diffusion model alongside text conditions. Inspired by prior adapter-based methods~\cite{mou2024t2i, ye2023ip}, GLIP-Adapter enables image-to-image translation guided by example-based visual conditioning.

We design GLIP-Adapter to leverage two complementary sources of information: global semantic context and local structural detail. Specifically, we extract:

\begin{itemize}
    \item \textbf{Global tokens} from CLIP~\cite{radford2021learning}, which capture high-level semantic content (e.g., scene category, weather type) based on joint image-text alignment.
    \item \textbf{Local tokens} from DINOv2~\cite{oquab2023dinov2}, which encode fine-grained spatial features such as object shape, boundary, and arrangement via self-supervised learning.
\end{itemize}

These global and local tokens are concatenated and projected via a lightweight adapter network, which is trained while keeping the image encoders and diffusion backbone frozen. The resulting prompt embedding $c_i$ is injected into the denoising model through cross-attention layers, alongside the text prompt $c_t$. The modified loss function becomes:

\begin{equation}
    \mathcal{L} = \mathbb{E}_{x_0, \epsilon \sim \mathcal{N}(0, I), c_t, c_i, t} \left\| \epsilon - \epsilon_\theta(x_t, c_t, c_i, t) \right\|^2,
    \label{eq:glip_loss}
\end{equation}

By combining CLIP's global semantic awareness with DINOv2's fine-grained spatial representation, GLIP-Adapter enables the model to maintain the layout, structure, and visual identity of the input image throughout the translation process. This design improves upon prior prompt-based methods (e.g., IP-Adapter) that rely solely on global-level image embeddings and often fail to preserve object-level consistency in multi-domain scenarios. To support downstream cross-attention operations such as those in our Multi-Domain Control Guidance (MCG), we apply a linear projection to align the prompt embeddings to a unified feature dimension.


\subsection{Multi-Domain Control Guidance (MCG)}

To enable flexible and interpretable control over domain-specific attribute translation, we propose Multi-Domain Control Guidance (MCG), a prompt-driven conditioning mechanism that operates on the difference between noise predictions guided by source and target prompts. Unlike traditional classifier-free guidance (CFG)~\cite{ho2022classifier}, which interpolates between unconditional and conditional predictions, MCG explicitly models the domain shift between attributes in the noise space.

Given a source domain prompt $r$ (e.g., \textit{cloudy}) and a target domain prompt $\hat{r}$ (e.g., \textit{sunny}), the diffusion model computes two conditional noise predictions:

\[
\epsilon_\theta(x_t, r, c_i, t) \quad \text{and} \quad \epsilon_\theta(x_t, \hat{r}, c_i, t),
\]

where $c_i$ is the image prompt from GLIP-Adapter. The translation direction is derived from the difference between these two predictions. The final guided noise is obtained by adding a scaled difference to the source-prompt prediction:

\begin{multline}
    \hat{\epsilon}_\theta(x_t, \hat{r}, c_i, t) =
    \epsilon_\theta(x_t, r, c_i, t) + \\
    s \cdot \left(
    \epsilon_\theta(x_t, \hat{r}, c_i, t) -
    \epsilon_\theta(x_t, r, c_i, t)
    \right),
    \label{eq:mcg_single}
\end{multline}

where $s$ is a translation scale parameter that controls the strength of attribute change. This formulation enables the model to retain source characteristics and apply only the directional change specified by the prompt difference.

\input{table/table_experiment_1_multi_domain_translation_celeba}

\input{table/table_experiment_2_multi_domain_translation_bdd100k}


To extend this mechanism to multi-domain translation, we introduce compositional control via linear combination. Let $\{(r_d, \hat{r}_d)\}_{d=1}^D$ be a set of $D$ source-target domain prompt pairs (e.g., \textit{cloudy} $\rightarrow$ \textit{sunny}, \textit{day} $\rightarrow$ \textit{night}). We apply independent noise deltas for each attribute dimension and scale them with domain-specific strengths $\{s_d\}_{d=1}^D$:

\begin{multline}
    \hat{\epsilon}_\theta(x_t, \hat{r}_d\, c_i, t) =
    \epsilon_\theta(x_t, r, c_i, t) +\\
    \sum_{d=1}^{D} s_d \cdot \left(
    \epsilon_\theta(x_t, \hat{r}_d, c_i, t) -
    \epsilon_\theta(x_t, r, c_i, t)
    \right).
    \label{eq:mcg_multi}
\end{multline}

This design enables three core capabilities:
\begin{itemize}
    \item \textbf{Selective translation:} Only attributes with differing source-target prompts are modified.
    \item \textbf{Compositional control:} Multiple attributes (e.g., weather, time, style) can be changed simultaneously via prompt composition.
    \item \textbf{Per-attribute scaling:} Each domain axis can be independently modulated using $s_d$, allowing fine-grained control over translation intensity.
\end{itemize}

Unlike prior approaches that rely on binary class labels or segmentation masks to isolate attributes, MCG enables soft, interpretable guidance using domain-aware prompt differences. Combined with GLIP-Adapter, this module provides a unified mechanism for structure-preserving, multi-domain translation with precise attribute-level control.


\section{Experiments}
\label{sec:experiments}

We evaluated our model on two real-world datasets, CelebA(Dialog) and BDD100K, each featuring multiple domain attributes such as facial expression, identity, weather, time, and scene type. Our experiments are designed to assess the model’s ability to (1) perform multi-attribute translations while preserving structural integrity, (2) enable compositional control across multiple domains, and (3) support fine-grained per-domain strength adjustment.

\input{figure/figure_1_domain_translation_celeba}

\input{figure/figure_2_domain_translation_bdd100k}

\subsection{Implementation Details}
Our method is implemented using the HuggingFace Diffusers library~\cite{von-platen-etal-2022-diffusers}, with Stable Diffusion v2.1 as the base model. During training, all images are resized to $512 \times 512$ and encoded into a latent space using the pretrained VAE. The model is fine-tuned using two NVIDIA A6000 GPUs for 200{,}000 steps with a batch size of 24 per GPU and a fixed learning rate of $1 \times 10^{-5}$. The adapters were trained using two A6000 GPUs for 200{,}000 steps, with a batch size of 24 per GPU, using CLIP-ViT-H/14~\cite{ilharco_gabriel_2021_5143773} and DINOv2-Large~\cite{oquab2023dinov2} as image encoders. For compatibility with baseline models, we utilized the HuggingFace Diffusers library~\cite{von-platen-etal-2022-diffusers} for all diffusion-based experiments. We also conducted additional experiment on animal face dataset to compare with existing text-guided translation models, which are described in the supplementary material. 

\subsection{Multi-Domain Image to Image Translation}
We conducted comparative experiments with existing multi-domain I2I translation models to evaluate the effectiveness of our method. For baseline models, we included SDEdit~\cite{meng2021sdedit}, DDIM, Direct Inversion~\cite{song2020denoising, ju2023direct} with editing methods MasaCtrl~\cite{cao2023masactrl}, Plug-and-Play~\cite{tumanyan2023plug}, and IP-Adapter~\cite{ye2023ip} to compare performance across a diverse range of models. Leveraging the multiple domain characteristics of our dataset, we conducted model performance evaluations by adjusting the number of translated domains. From 500 validation scenarios, we randomly selected one image and applied translations across a randomly chosen set of 1, 2 or 3 domains. \Cref{tab:multi_domain_translation_celeba},\Cref{tab:multi_domain_translation_bdd100k} show the results, indicating that our method achieved sota in most metrics,  or second-best performance across most evaluation metrics. This suggests that our model not only preserves the structural and contextual content of the source image, but also performs accurate and effective multi-domain translation. 

Furthermore, we observed that standard metrics such as FID and CLIP similarity may not fully capture the increasing difficulty of multi-attribute editing, as they often reward conservative, under-edited results that maintain high visual stability but fail to accurately reflect semantic changes. To address this limitation and better assess the true effectiveness of multi-domain translation, we conducted a human evaluation focusing on attribute correctness and visual naturalness. Five human evaluators assessed 100 samples from CelebA, BDD100K dataset. The evaluators rated the degree of semantic alignment between the visual outputs and the text prompts on a scale of 0 to 10, which were subsequently normalized to a range of $[0, 1]$. As shown in the Human column of \Cref{tab:multi_domain_translation_celeba} and \Cref{tab:multi_domain_translation_bdd100k}, while the performance of most baselines degrades significantly as the number of edited domains increases. For instance, IP-Adapter's score drops from 0.88 to 0.52, LACE maintains a nearly constant high level of correctness (from 0.92 to 0.90). These results confirm that LACE is robust in accurately applying all requested attributes, a characteristic that human evaluation highlights even when standard metrics remain relatively stable.

Qualitative evaluation was performed on four popular criteria: image quality (FID~\cite{heusel2017fid}, FID\textsubscript{clip}\cite{kynkaanniemi2022fidclip}), structure distance\cite{tumanyan2022splicing}, background preservation (PSNR, LPIPS~\cite{zhang2018lpips}, MSE, SSIM~\cite{wang2004ssim}), translation quality (CLIP Similarity~\cite{hessel2021clipscore}) to measure how well the visual outputs align with the linguistic semantics of the target prompt.
 Qualitative results in \Cref{fig:domain_translation_celeba},\Cref{fig:domain_translation_bdd100k} further demonstrate that our method produces visually coherent and attribute-consistent outputs across diverse translation scenarios.


\input{figure/figure_3_ablation_celeba}
\subsection{Ablation Study}

\input{table/table_experiment_3_ablation}

We conduct a series of ablation studies to evaluate the contribution of each component in our framework. The baseline is IP-Adapter~\cite{ye2023ip}, which uses CLIP-based global visual prompts. We progressively introduce modifications and measure performance changes across  3 domain translation settings.

\begin{itemize}
    \item \textbf{CLIP-Full:} We first enhance the baseline by incorporating patch-wise local tokens extracted from the CLIP encoder, in addition to the original global token. This modification improves both FID and FID-CLIP scores, indicating better image quality and structural preservation.
    
    \item \textbf{DINOv2-Full:} Next, we replace CLIP with DINOv2 as the visual encoder, which provides stronger spatial features. This leads to further improvements in structure-sensitive metrics and better preservation of Local-level details such as object color and shape visualized in \cref{fig:ablation}.
    
    \item \textbf{GLIP-Adapter:}  GLIP-Adapter builds on DINOv2-Full by replacing the global token with that of CLIP while retaining the local tokens from DINOv2.
    \item \textbf{MCG:} We compare our Multi-Domain Control Guidance (MCG) with traditional Classifier-Free Guidance (CFG)~\cite{ho2022classifier}. While CFG interpolates between unconditional and conditional denoising, MCG explicitly models attribute shift using the difference between source and target prompt predictions. Our experiments show substantial performance gains in FID and CLIP similarity when using MCG.
    
    \item \textbf{DDIM:} Finally, we incorporate DDIM Inversion~\cite{song2020denoising} to accurately reconstruct the initial noise from the input image. This further enhances the preservation of structural and contextual details in complex scenes.
\end{itemize}

Quantitative results are reported in \Cref{tab:ablation}, and visual comparisons are shown in \cref{fig:ablation}, both demonstrating the effectiveness of each proposed component in improving translation fidelity, structural consistency, and controllability.

\subsection{Translation Scale Control}
We demonstrate the model’s capability to adjust the strength of domain translation via the scaling factor $s$ on the CelebA dataset. As shown in \cref{fig:scale_single_celeba}, increasing $s$ amplifies the degree of stylistic transfer from the target domain while maintaining the structural integrity of the source image. Lower values of $s$ result in subtle modifications, preserving more of the original domain’s appearance, whereas higher values push the model to emphasize features more closely aligned with the target domain. This behavior validates the model's ability to perform controllable and progressive translation, which is particularly important for applications requiring user-interactive manipulation or gradual domain adaptation. The use of a continuous scalar $s$ introduces a simple yet effective mechanism to traverse the interpolation space between source and target domains without retraining the model.

\subsection{Per-Domain Scaling}
While global scaling via a single $s$ value is effective for coarse control, many real-world applications demand finer, attribute-specific adjustments, especially in complex multi-domain translation settings. To address this, we introduce a differential translation scaling mechanism by assigning separate scaling factors $s_1, s_2, \dots, s_D$ for each of the $D$ domain attributes. On the BDD100K dataset, we apply distinct scaling coefficients to individual domain factors such as weather and time-of-day. As illustrated in \cref{fig:scale_multi_bdd100k}, this allows precise, independent modulation of each attribute's transformation strength, enabling complex yet interpretable compositions, for instance, increasing snowfall while keeping the scene at dusk. Such capability facilitates tailored image generation that adapts to specific user intents or context-aware conditions.

Unlike traditional methods that use a single unified scaling parameter, our approach disentangles the contribution of each domain axis, allowing diverse combinations of partial translations within a single forward pass. This fine-grained translation control significantly expands the model's expressiveness.

\section{Conclusion}
\label{sec:conclusion}

We presented LACE, a diffusion-based framework for controllable multi-domain image-to-image translation, addressing both structural preservation and fine-grained attribute control. Our approach introduces two key components:(1) the Global-Local Image Prompt Adapter (GLIP-Adapter), which fuses semantic and spatial cues from the input image to guide structure-aware translation, and(2) the Multi-Domain Control Guidance (MCG), which enables targeted and compositional editing via prompt-driven attribute steering.
Unlike prior methods that rely on spatial masks or support only single-attribute editing, our framework enables flexible and interpretable translation across multiple domain axes without compromising scene consistency.
Experiments on CelebA and BDD100K demonstrate superior performance in visual fidelity, structural integrity, and multi-attribute controllability. Further, ablation and scale control studies confirming that controllable image translation can be framed as a language grounding problem, where semantic differences between prompts drive structured, domain-specific transformations.

\section*{Limitations}

While LACE demonstrates strong controllability and structural preservation across both fine-grained attribute editing (CelebA) and domain-level translation (BDD100K), several limitations remain.

First, the framework introduces additional computational overhead: the GLIP-Adapter and multi-domain guidance require multiple noise predictions per domain, and inference cost grows as more attributes are edited simultaneously, limiting real-time applicability.

Second, our evaluation is restricted to three datasets (CelebA, BDD100K, Animal Faces). Although these cover both local attributes and global style/background domains, broader validation on more diverse domains (e.g., medical images, artwork, video) is left for future work.

Third, attribute interference may arise when editing multiple factors simultaneously: While per-domain scaling alleviates this issue, semantic conflicts between attributes (e.g., gender and hair length) remain a challenge.

Fourth, the framework depends on source–target prompt differences, making it sensitive to linguistic ambiguity; future work could incorporate advances in natural language understanding or prompt paraphrasing to improve robustness; better prompt robustness or automatic prompt refinement could further improve stability.

\section*{Ethical Considerations}
As a controllable image translation framework, LACE could be misused for generating or altering visual content in deceptive ways. 
While our work is intended for research and data augmentation under ethical use, future extensions should include watermarking or misuse detection mechanisms to mitigate potential risks.

\section*{Acknowledgements}

This work was partly supported by the Technology development Program [RS-2025-25443237] funded by the Ministry of SMEs and Startups(MSS, Korea) and partly supported by the Institute of Information \&
Communications Technology Planning \& Evaluation (IITP) grant funded by the
Korea government (MSIT) [RS-2021-II211341, Artificial Intelligence Graduate
School Program (Chung-Ang University)].

\bibliography{custom}

\appendix

\section{Appendix}
\input{figure/figure_9_animal_face}

\input{figure/figure_5_scale_single_celeba}
\input{figure/figure_8_scale_multi_bdd100k}

\end{document}

%% file: table/table_experiment_1_multi_domain_translation_celeba.tex
\begin{table*}[t]
    \centering
    \resizebox{\textwidth}{!}{
    \begin{tabular}{c|l|ccc|cccc|c|c}
    \toprule
        \multirow{2}*{\shortstack{Number of\\ translations}}  & \multirow{2}*{Methods}  & \multirow{2}*{FID$\downarrow$} & \multirow{2}*{FID\textsubscript{clip}$\downarrow$} & \multirow{2}*{\shortstack{Structure\\Distance$\downarrow$}} & \multicolumn{4}{c|}{Background Preservation} & \multirow{2}*{\shortstack{CLIP\\Similarity$\uparrow$}} & \multirow{2}*{Human$\uparrow$}  \\ \cline{6-9} 
                            &                 &       &           &  & \multicolumn{1}{c}{PSNR$\uparrow$} & \multicolumn{1}{c}{LPIPS$\downarrow$} & \multicolumn{1}{c}{MSE$\downarrow$} & SSIM$\uparrow$ &  &  \\ \hline
        \multirow{7}* [-.3ex]{1 domain}

        & SDEdit~\cite{meng2021sdedit}          & 53.09 & 15.59 & 0.1374 & \multicolumn{1}{c}{13.73} & \multicolumn{1}{c}{0.49} & \multicolumn{1}{c}{0.0278} & 0.58 & 21.74 & 0.78 \\

                                         & DDIM~\cite{song2020denoising}+PnP~\cite{tumanyan2023plug}       & 69.10 & 25.00 & 0.1455 & \multicolumn{1}{c}{15.01} & \multicolumn{1}{c}{0.40} & \multicolumn{1}{c}{0.0273} & 0.60 & \underline{22.04} & 0.82 \\ 
                                         
                                         & DDIM~\cite{song2020denoising}+MasaCtrl~\cite{cao2023masactrl}   & 164.74 & 53.96 & 0.1747 & \multicolumn{1}{c}{\underline{15.58}} & \multicolumn{1}{c}{0.49} & \multicolumn{1}{c}{0.0962} & 0.60 & 15.83 & 0.65 \\ 
                                         
                                         & Direct~\cite{ju2023direct}+PnP~\cite{tumanyan2023plug}      & 60.52 & 15.59 & 0.1195 & \multicolumn{1}{c}{15.34} & \multicolumn{1}{c}{0.37} & \multicolumn{1}{c}{0.0438} & \textbf{0.77} & 20.31 & 0.81 \\

                                         & Direct~\cite{ju2023direct}+MasaCtrl~\cite{cao2023masactrl} & 49.31 & 17.29 & 0.0854 & \multicolumn{1}{c}{10.75} & \multicolumn{1}{c}{0.36} & \multicolumn{1}{c}{0.0336} & 0.68 & 17.29 & 0.79 \\

                                         & IP-Adapter~\cite{ye2023ip}     & \underline{47.53} &    \underline{11.38}    & \underline{0.0629}    & \multicolumn{1}{c}{15.34 } & \multicolumn{1}{c}{ \underline{0.32}} & \multicolumn{1}{c}{\underline{0.0257}}  &  0.56 &  17.73 & \underline{0.88} \\ \cline{2-11}

                                         & LACE (Ours) &\textbf{ 45.50} & \textbf{10.61} & \textbf{0.0622} & \multicolumn{1}{c}{\textbf{16.24}} & \multicolumn{1}{c}{\textbf{0.30}} & \multicolumn{1}{c}{\textbf{0.0252}} & \underline{0.73} & \textbf{23.12} & \textbf{0.91} \\ \hline \hline
        \multirow{7}* [-.3ex]{2 domains} 
        
        & SDEdit~\cite{meng2021sdedit}          & 56.97 & 18.31 & 0.1465 & \multicolumn{1}{c}{14.12} & \multicolumn{1}{c}{0.52} & \multicolumn{1}{c}{0.0359} & 0.55 & 22.35 & 0.61 \\
        
                                         & DDIM~\cite{song2020denoising}+PnP~\cite{tumanyan2023plug}         & 79.53 & 28.92 & 0.1533 & \multicolumn{1}{c}{15.97} & \multicolumn{1}{c}{0.41} & \multicolumn{1}{c}{0.0342} & 0.57 & \underline{22.70} & 0.68 \\ 
                                         
                                         & DDIM~\cite{song2020denoising}+MasaCtrl~\cite{cao2023masactrl}   & 159.98 & 54.01 & 0.1820 & \multicolumn{1}{c}{15.40} & \multicolumn{1}{c}{0.49} & \multicolumn{1}{c}{0.0983} & 0.59 & 16.56 & 0.48 \\
                                         
                                         & Direct~\cite{ju2023direct}+PnP~\cite{tumanyan2023plug}      & 70.03 & 19.38 & 0.1175 & \multicolumn{1}{c}{11.89} & \multicolumn{1}{c}{0.41} & \multicolumn{1}{c}{0.0473} & \textbf{0.73} & 20.90 & 0.66 \\
                                         
                                         & Direct~\cite{ju2023direct}+MasaCtrl~\cite{cao2023masactrl} & 56.12 & 19.88 & 0.0981 & \multicolumn{1}{c}{10.55} & \multicolumn{1}{c}{0.39} & \multicolumn{1}{c}{0.0381} & 0.64 & 16.09 & 0.63 \\
                                         
                                         & IP-Adapter~\cite{ye2023ip}       &  \underline{46.57} &  \underline{13.45}     &  \underline{0.0783}  & \multicolumn{1}{c}{\underline{16.35}} & \multicolumn{1}{c}{\underline{0.38} } & \multicolumn{1}{c}{ \textbf{0.0315}} & 0.57 &   17.87 & \underline{0.70}
                                         \\ \cline{2-11}

                                         & LACE (Ours) & \textbf{45.77} & \textbf{11.98 } & \textbf{0.0761} & \multicolumn{1}{c}{\textbf{16.90}} & \multicolumn{1}{c}{\textbf{0.33}} & \multicolumn{1}{c}{\underline{0.0340}} &\underline{0.69} & \textbf{23.38} & \textbf{0.92} \\ \hline \hline

        \multirow{7}* [-.3ex]{3 domains}

        & SDEdit~\cite{meng2021sdedit}         & 58.97 & 19.74 & 0.1485 & \multicolumn{1}{c}{13.85} & \multicolumn{1}{c}{0.53} & \multicolumn{1}{c}{0.0398} & 0.54 & 22.27 & 0.45 \\

                                         & DDIM~\cite{song2020denoising}+PnP~\cite{tumanyan2023plug}         & 76.28 & 28.65 & 0.1553 & \multicolumn{1}{c}{15.75} & \multicolumn{1}{c}{0.41} & \multicolumn{1}{c}{0.0397} & 0.55 & \underline{22.60} & 0.54 \\

                                         & DDIM~\cite{song2020denoising}+MasaCtrl~\cite{cao2023masactrl}   & 158.40 & 53.86 & 0.1864 & \multicolumn{1}{c}{15.26} & \multicolumn{1}{c}{0.50} & \multicolumn{1}{c}{0.1009} & 0.58 & 16.83 & 0.32 \\

                                         & Direct~\cite{ju2023direct}+PnP~\cite{tumanyan2023plug}      & 73.93 & 21.71 & 0.2219 & \multicolumn{1}{c}{10.58} & \multicolumn{1}{c}{0.43} & \multicolumn{1}{c}{0.0495} & \textbf{0.70} & 21.22 & 0.51 \\

                                         & Direct~\cite{ju2023direct}+MasaCtrl~\cite{cao2023masactrl} & 54.94 & 22.49 & 0.1168 & \multicolumn{1}{c}{10.86} & \multicolumn{1}{c}{\underline{0.42}} & \multicolumn{1}{c}{0.0451} & 0.60 & 14.01 & 0.48 \\

                                         & IP-Adapter~\cite{ye2023ip}       & \underline{47.56}  &   \underline{15.84}      &  \underline{0.0957}   & \multicolumn{1}{c}{ \underline{16.28}} & \multicolumn{1}{c}{ 0.47} & \multicolumn{1}{c}{\underline{0.0392} } & 0.55  & 17.62  & \underline{0.52} \\ \cline{2-11}

                                         & LACE (Ours) & \textbf{46.17} & \textbf{13.32} & \textbf{0.0833} & \multicolumn{1}{c}{\textbf{16.31}} & \multicolumn{1}{c}{\textbf{0.34}} & \multicolumn{1}{c}{\textbf{0.0388}} &\underline{ 0.68} & \textbf{23.24} & \textbf{0.92} \\
    \bottomrule
    \end{tabular}
    }
    \vspace{-5pt}
    \caption {\textbf{Quantitative evaluation for multi-domain image-to-image translation methods on CelebA}. The best results are highlighted in bold, the second best results are marked with an underline.}
    \label{tab:multi_domain_translation_celeba}
    \vspace{-10pt}
\end{table*}

%% file: table/table_experiment_2_multi_domain_translation_bdd100k.tex
\begin{table*}[t]
    \centering
    \resizebox{\textwidth}{!}{
    \begin{tabular}{c|l|ccc|cccc|c|c}
    \toprule
        \multirow{2}*{\shortstack{Number of\\ translations}}  & \multirow{2}*{Methods}  & \multirow{2}*{FID$\downarrow$} & \multirow{2}*{FID\textsubscript{clip}$\downarrow$} & \multirow{2}*{\shortstack{Structure\\Distance$\downarrow$}} & \multicolumn{4}{c|}{Background Preservation} & \multirow{2}*{\shortstack{CLIP\\Similarity$\uparrow$}}& \multirow{2}*{Human$\uparrow$}  \\ \cline{6-9} 
                            &                 &       &           &  & \multicolumn{1}{c}{PSNR$\uparrow$} & \multicolumn{1}{c}{LPIPS$\downarrow$} & \multicolumn{1}{c}{MSE$\downarrow$} & SSIM$\uparrow$ &  &  \\ \hline
        \multirow{7}* [-.3ex]{1 domain}

        & SDEdit~\cite{meng2021sdedit}          & 42.95 & 11.50 & \textbf{0.0389} & \multicolumn{1}{c}{\underline{20.87}} & \multicolumn{1}{c}{\underline{0.23}} & \multicolumn{1}{c}{\underline{0.0157}} & \underline{0.66} & 17.99 & 0.75 \\

                                         & DDIM~\cite{song2020denoising}+PnP~\cite{tumanyan2023plug}      & 48.36 & 8.49 & 0.0737 & \multicolumn{1}{c}{14.80} & \multicolumn{1}{c}{0.34} & \multicolumn{1}{c}{0.0392} & 0.53 & \underline{18.68} & 0.81 \\

                                         & DDIM~\cite{song2020denoising}+MasaCtrl~\cite{cao2023masactrl}   & 112.53 & 27.52 & 0.1196 & \multicolumn{1}{c}{12.22} & \multicolumn{1}{c}{0.43} & \multicolumn{1}{c}{0.0664} & 0.43 & 15.37 & 0.62 \\

                                         & Direct~\cite{ju2023direct}+PnP~\cite{tumanyan2023plug}      & \underline{40.56} & \textbf{6.11} & 0.0719 & \multicolumn{1}{c}{15.13} & \multicolumn{1}{c}{0.31} & \multicolumn{1}{c}{0.0372} & 0.54 & 18.29 & 0.83 \\

                                         & Direct~\cite{ju2023direct}+MasaCtrl~\cite{cao2023masactrl} & 73.44 & 15.75 & 0.0959 & \multicolumn{1}{c}{13.59} & \multicolumn{1}{c}{0.38} & \multicolumn{1}{c}{0.0531} & 0.49 & 16.00 & 0.79 \\

                                         & IP-Adapter~\cite{ye2023ip}      &  54.05 &   12.59   & 0.065     & \multicolumn{1}{c}{17.34 } & \multicolumn{1}{c}{ 0.32} & \multicolumn{1}{c}{ 0.022} & 0.56  & 18.71  & \underline{0.88} \\ \cline{2-11}

                                         & LACE (Ours) &  \textbf{40.15}  &  \underline{ 7.53}   &    \underline{0.0453} & \multicolumn{1}{c}{\textbf{21.96} } & \multicolumn{1}{c}{\textbf{0.21} } & \multicolumn{1}{c}{\textbf{0.0092} } & \textbf{0.73} & \textbf{18.82}  & \textbf{0.91} \\ \hline \hline
        \multirow{7}* [-.3ex]{2 domains}

        & SDEdit~\cite{meng2021sdedit}          & 44.53 & 11.92 & \textbf{0.0389} & \multicolumn{1}{c}{\underline{20.87}} & \multicolumn{1}{c}{\underline{0.23}} & \multicolumn{1}{c}{\underline{0.0099}} &\underline{ 0.66} & 16.04 & 0.58 \\

                                         & DDIM~\cite{song2020denoising}+PnP~\cite{tumanyan2023plug}        & 49.62 & 9.15 & 0.0752 & \multicolumn{1}{c}{14.93} & \multicolumn{1}{c}{0.34} & \multicolumn{1}{c}{0.0378} & 0.54 & \underline{17.29} & 0.64 \\

                                         & DDIM~\cite{song2020denoising}+MasaCtrl~\cite{cao2023masactrl}   & 113.84 & 27.54 & 0.1213 & \multicolumn{1}{c}{12.25} & \multicolumn{1}{c}{0.43} & \multicolumn{1}{c}{0.0663} & 0.43 & 14.85 & 0.45 \\

                                         & Direct~\cite{ju2023direct}+PnP~\cite{tumanyan2023plug}      & \underline{43.83} & \underline{7.07} & 0.0736 & \multicolumn{1}{c}{15.21} & \multicolumn{1}{c}{0.31} & \multicolumn{1}{c}{0.0367} & 0.55 & 16.88 & 0.67 \\

                                         & Direct~\cite{ju2023direct}+MasaCtrl~\cite{cao2023masactrl} & 63.53 & 13.61 & 0.0932 & \multicolumn{1}{c}{13.61} & \multicolumn{1}{c}{0.37} & \multicolumn{1}{c}{0.0499} & 0.49 & 13.86 & 0.61 \\

                                         & IP-Adapter~\cite{ye2023ip}       &  54.72 &   12.68     & 0.0652   & \multicolumn{1}{c}{ 17.35} & \multicolumn{1}{c}{0.32 } & \multicolumn{1}{c}{0.022 } & 0.56&  16.55 & \underline{0.70} \\ \cline{2-11}
                                         
                                         & LACE (Ours) &\textbf{ 40.53}  &      \textbf{7.64}    &   \underline{0.0466}  & \multicolumn{1}{c}{\textbf{21.98} } & \multicolumn{1}{c}{\textbf{0.21} } & \multicolumn{1}{c}{ \textbf{0.0091}} &  \textbf{0.72} &  \textbf{18.80} & \textbf{0.90} \\ \hline \hline

        \multirow{7}* [-.3ex]{3 domains}
        
        & SDEdit~\cite{meng2021sdedit}          & \underline{ 45.61} & 11.87       & \textbf{0.039}     & \multicolumn{1}{c}{\underline{20.87}} & \multicolumn{1}{c}{\underline{0.23}} & \multicolumn{1}{c}{\underline{0.0132}} & \underline{0.66} & 12.99 & 0.44 \\

                                         & DDIM~\cite{song2020denoising}+PnP~\cite{tumanyan2023plug}         & 51.00 & 9.71       & 0.0759     & \multicolumn{1}{c}{15.01} & \multicolumn{1}{c}{0.34} & \multicolumn{1}{c}{0.0372} & 0.53 & \underline{15.09} & \underline{0.54} \\

                                         & DDIM~\cite{song2020denoising}+MasaCtrl~\cite{cao2023masactrl}   & 115.11 & 27.97     & 0.1234     & \multicolumn{1}{c}{12.18} & \multicolumn{1}{c}{0.44} & \multicolumn{1}{c}{0.0673} & 0.43 & 12.39 & 0.31 \\

                                         & Direct~\cite{ju2023direct}+PnP~\cite{tumanyan2023plug}        & 46.61 & \underline{7.83}       & 0.075     & \multicolumn{1}{c}{15.25} & \multicolumn{1}{c}{0.31} & \multicolumn{1}{c}{0.0363} & 0.55 & 14.37 & 0.53 \\

                                         & Direct~\cite{ju2023direct}+MasaCtrl~\cite{cao2023masactrl} & 62.71 & 13.16       & 0.0905     & \multicolumn{1}{c}{13.69} & \multicolumn{1}{c}{0.37} & \multicolumn{1}{c}{0.0492} & 0.49 & 10.61 & 0.48 \\

                                         & IP-Adapter~\cite{ye2023ip}       & 55.09  &   12.88     & 0.0654     & \multicolumn{1}{c}{ 17.35} & \multicolumn{1}{c}{ 0.32} & \multicolumn{1}{c}{ 0.0219} &0.56 &  13.15 & 0.52 \\ \cline{2-11}
                                         
                                         & LACE (Ours) & \textbf{42.53}  &      \textbf{7.76}    &  \underline{ 0.0494}  & \multicolumn{1}{c}{\textbf{21.95} } & \multicolumn{1}{c}{\textbf{0.22} } & \multicolumn{1}{c}{\textbf{0.0106} } & \textbf{0.69}  &  \textbf{17.75} & \textbf{0.90} \\ 
    \bottomrule
    \end{tabular}
    }
    \vspace{-5pt}
    \caption{\textbf{Quantitative evaluation for multi-domain image-to-image translation methods on BDD100K}. The best results are highlighted in bold, the second best results are marked with an underline.}
    \label{tab:multi_domain_translation_bdd100k}
    \vspace{-15pt}
\end{table*}

%% file: figure/figure_1_domain_translation_celeba.tex
\begin{figure*}[t]
\centering
\resizebox{\textwidth}{!}{%
\begin{minipage}{\textwidth}
    \centering
    \makebox[0.03\textwidth]{}
    \makebox[0.15\textwidth]{\scriptsize Source}
    \makebox[0.15\textwidth]{\scriptsize SDEdit}
    \makebox[0.15\textwidth]{\scriptsize DDIM+MasaCtrl}
    \makebox[0.15\textwidth]{\scriptsize Direct+PnP}
    \makebox[0.15\textwidth]{\scriptsize IP-Adapter}
    \makebox[0.15\textwidth]{\scriptsize LACE (Ours)}
    \\
    \raisebox{0.2\height}{\makebox[0.03\textwidth]{\rotatebox{90}{\makecell{\scriptsize 1 domain}}}}
    \includegraphics[width=0.15\textwidth]{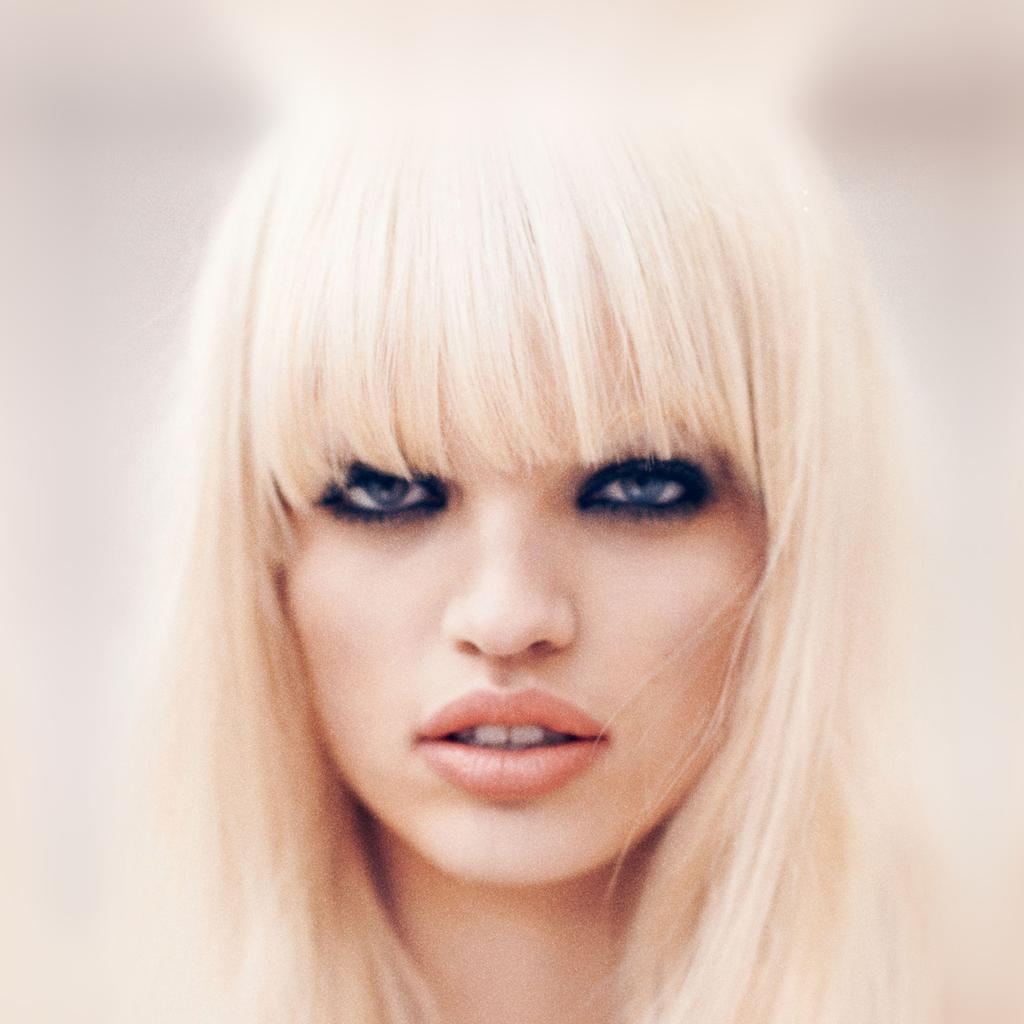}
    \includegraphics[width=0.15\textwidth]{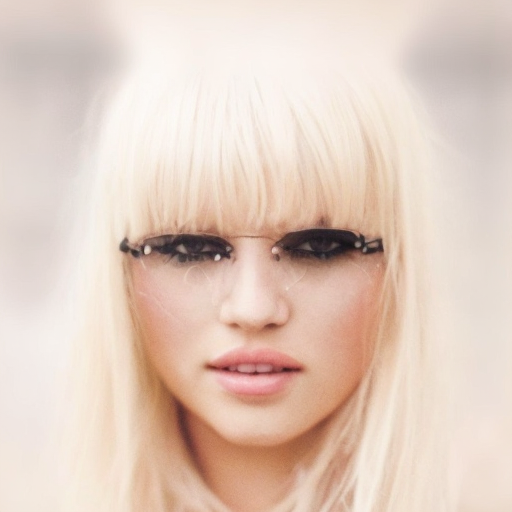}
    \includegraphics[width=0.15\textwidth]{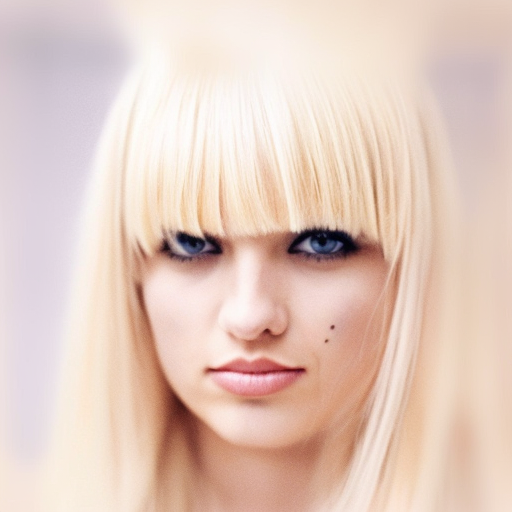}
    \includegraphics[width=0.15\textwidth]{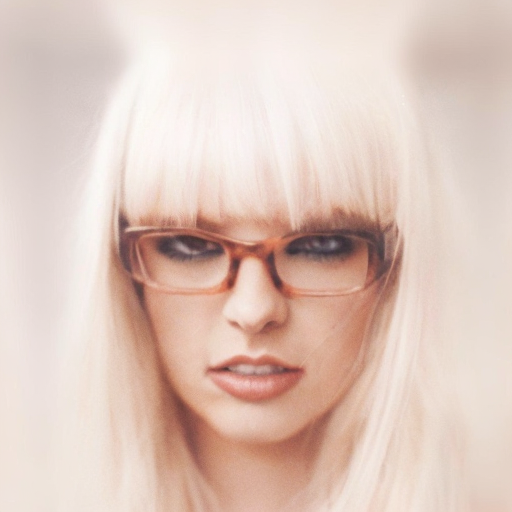}
    \includegraphics[width=0.15\textwidth]{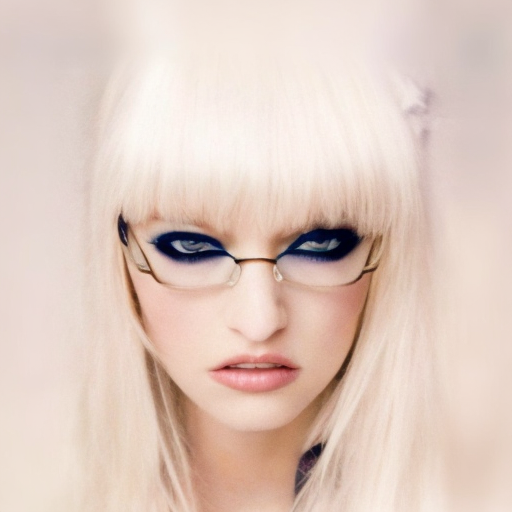}
    \includegraphics[width=0.15\textwidth]{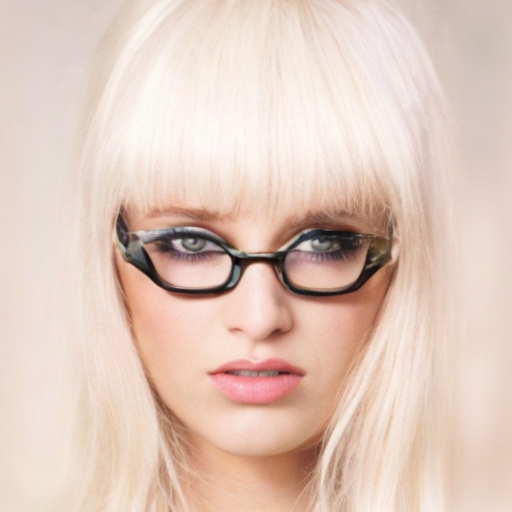}
    \\
    \raisebox{0.8\height}{\makebox[0.9\textwidth]{\scriptsize no smile / young / no glasses→\textcolor{blue}{glasses} / female/ extremely long bangs}}
    \\
    \raisebox{0.2\height}{\makebox[0.03\textwidth]{\rotatebox{90}{\makecell{\scriptsize 2 domain}}}}
    \includegraphics[width=0.15\textwidth]{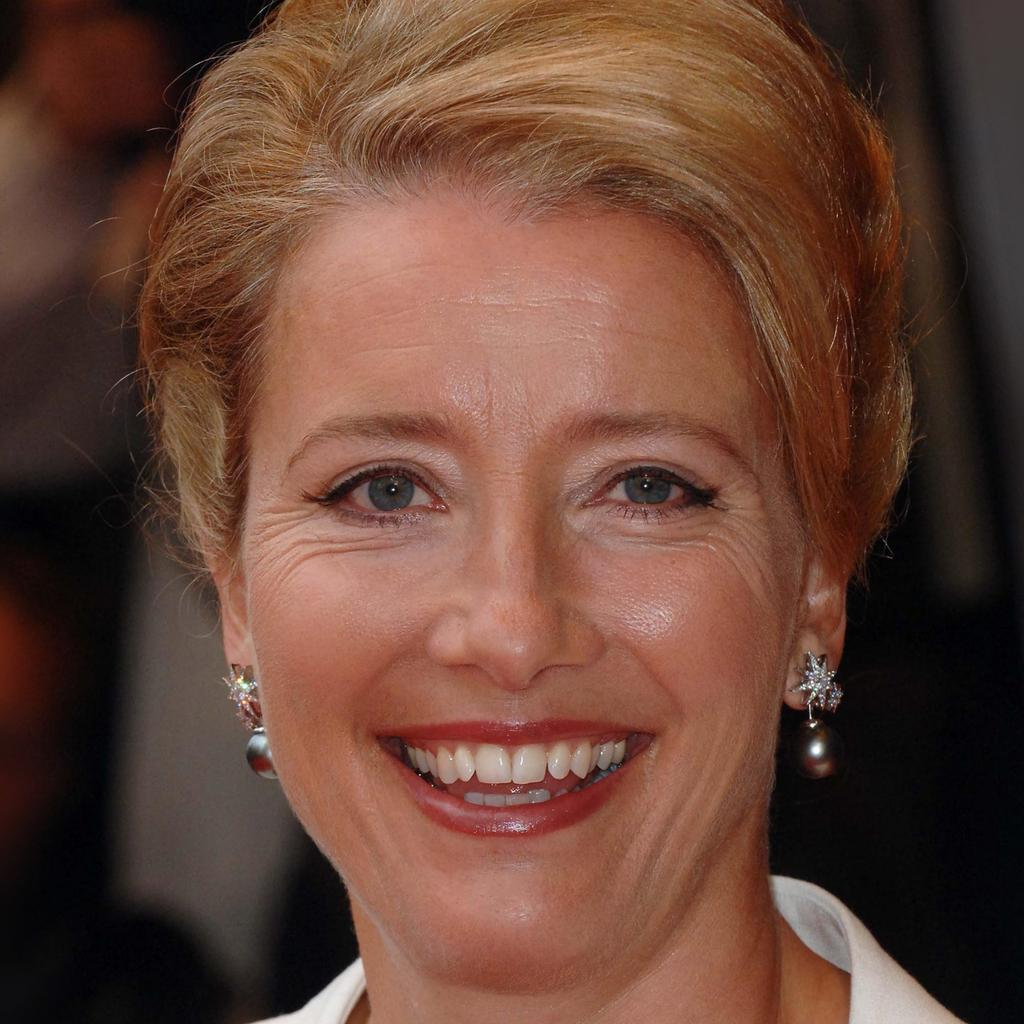}
    \includegraphics[width=0.15\textwidth]{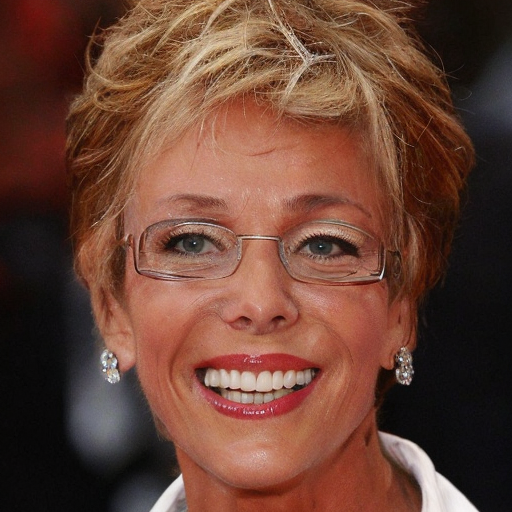}
    \includegraphics[width=0.15\textwidth]{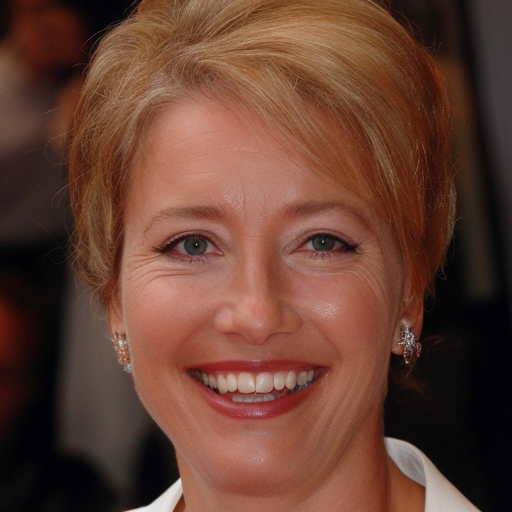}
    \includegraphics[width=0.15\textwidth]{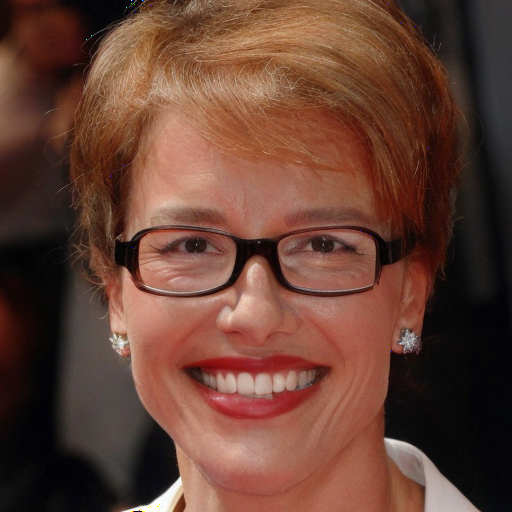}
    \includegraphics[width=0.15\textwidth]{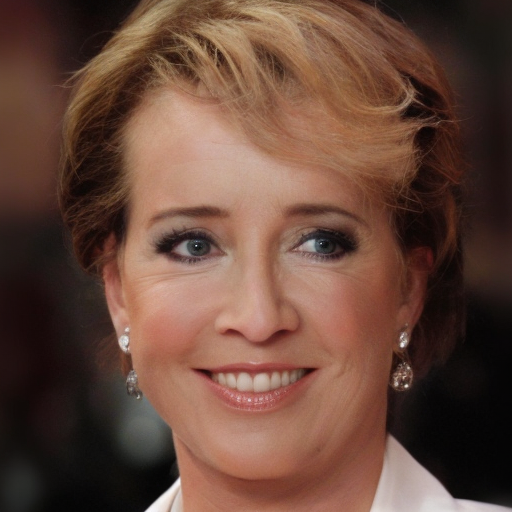}
    \includegraphics[width=0.15\textwidth]{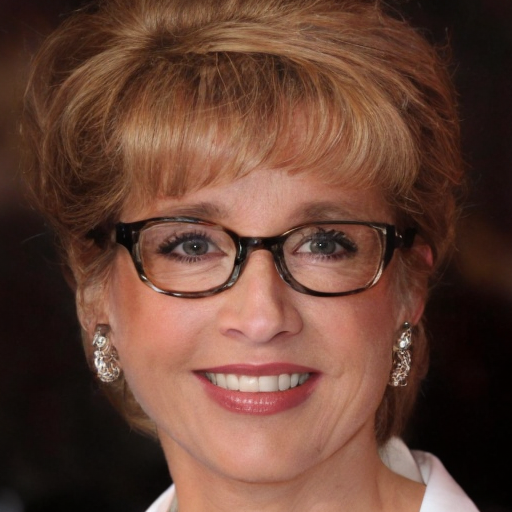}
    \\
    \raisebox{0.8\height}{\makebox[0.9\textwidth]{\scriptsize no eyeglasses→\textcolor{blue}{thin frame eyeglasses}  / no bangs→\textcolor{blue}{short bangs}  / beaming face / female / middle age}}
    \\
    \raisebox{0.2\height}{\makebox[0.03\textwidth]{\rotatebox{90}{\makecell{\scriptsize 3 domain}}}}
    \includegraphics[width=0.15\textwidth]{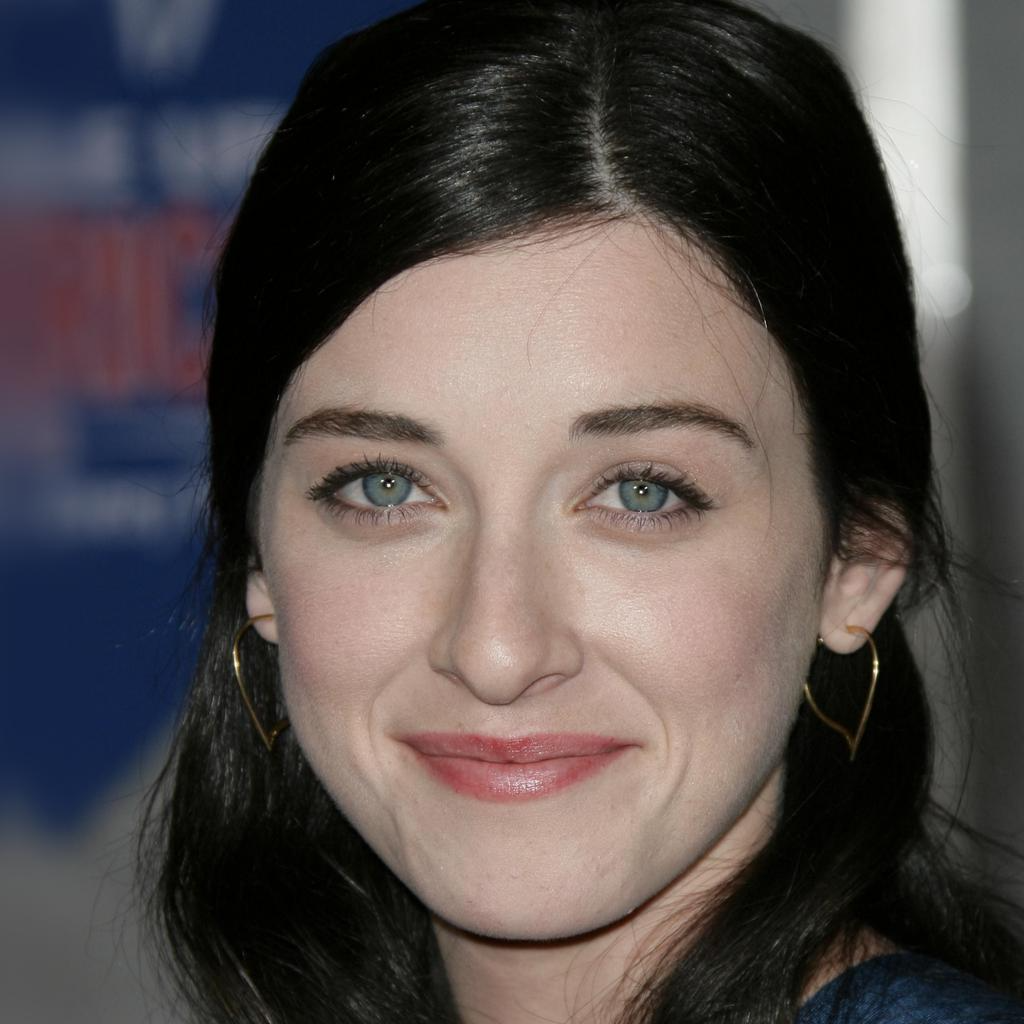}
    \includegraphics[width=0.15\textwidth]{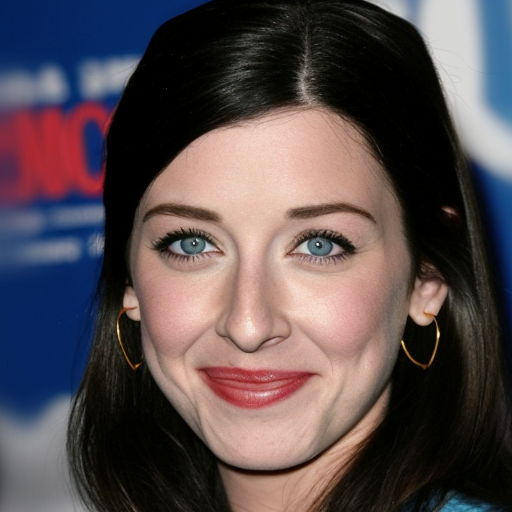}
    \includegraphics[width=0.15\textwidth]{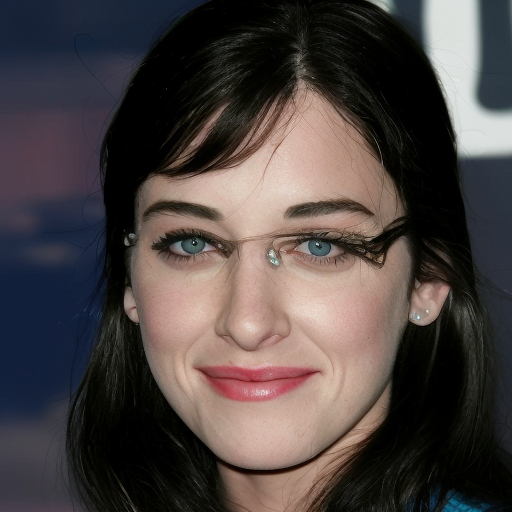}
    \includegraphics[width=0.15\textwidth]{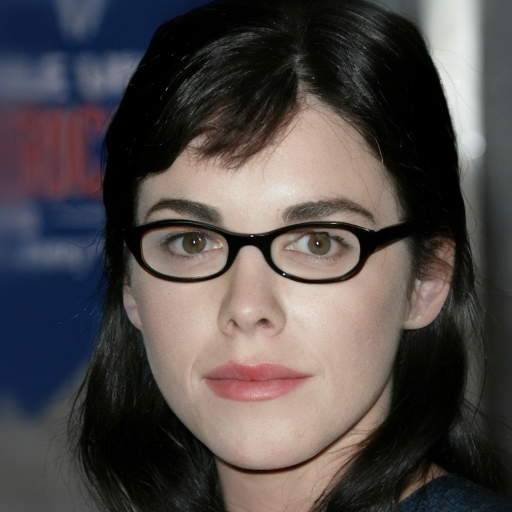}
    \includegraphics[width=0.15\textwidth]{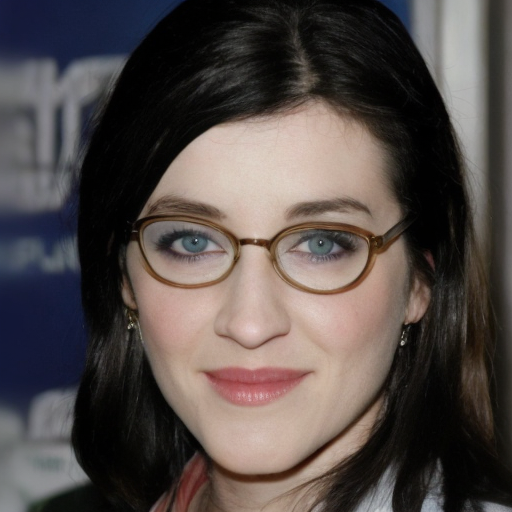}
    \includegraphics[width=0.15\textwidth]{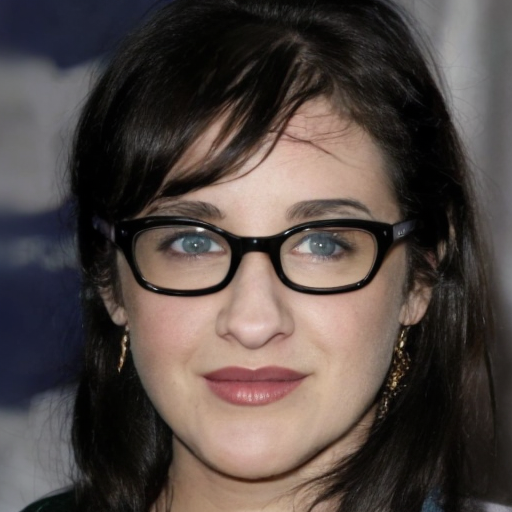}
    \\
    \raisebox{0.7\height}{\makebox[0.9\textwidth]{\scriptsize without eyeglasses→\textcolor{blue}{glasses}  / visible forehead→\textcolor{blue}{long bangs}  / smile→\textcolor{blue}{serious face} / female / young}}

\end{minipage}%
}
\vspace{-5pt}
\caption{\textbf{Qualitative evaluation for multi-domain image-to-image translation methods on CelebA.}}
\label{fig:domain_translation_celeba}
\vspace{-10pt}
\end{figure*}

%% file: figure/figure_2_domain_translation_bdd100k.tex
\begin{figure*}[t]
\centering
\resizebox{\textwidth}{!}{%
\begin{minipage}{\textwidth}
   \centering
    \makebox[0.03\textwidth]{}
    \makebox[0.15\textwidth]{\scriptsize Source}
    \makebox[0.15\textwidth]{\scriptsize SDEdit}
    \makebox[0.15\textwidth]{\scriptsize DDIM+MasaCtrl}
    \makebox[0.15\textwidth]{\scriptsize Direct+PnP}
    \makebox[0.15\textwidth]{\scriptsize IP-Adapter}
    \makebox[0.15\textwidth]{\scriptsize LACE (Ours)}
    \\
    \raisebox{0.2\height}{\makebox[0.03\textwidth]{\rotatebox{90}{\makecell{\scriptsize 1 domain}}}}
    \includegraphics[width=0.15\textwidth]{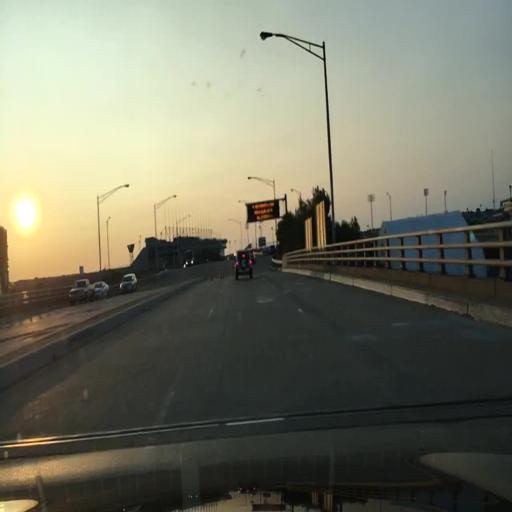}
    \includegraphics[width=0.15\textwidth]{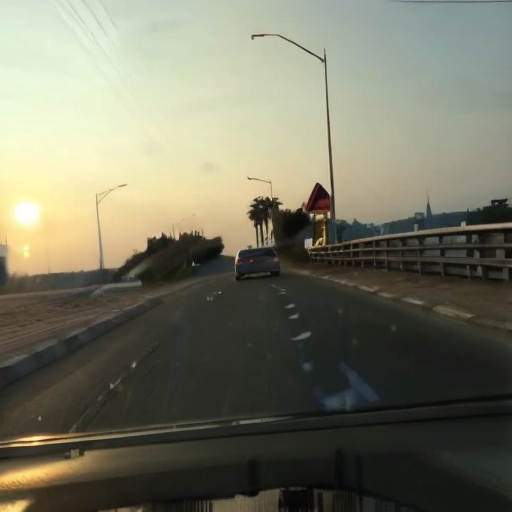}
    \includegraphics[width=0.15\textwidth]{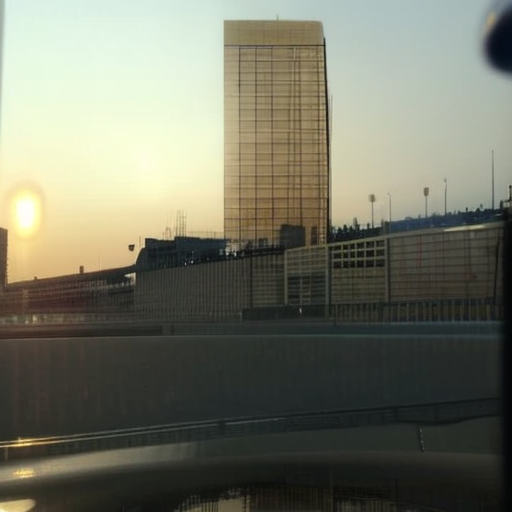}
    \includegraphics[width=0.15\textwidth]{images/domain_translation/bdd100k/1domain/direct+pnp.png}
    \includegraphics[width=0.15\textwidth]{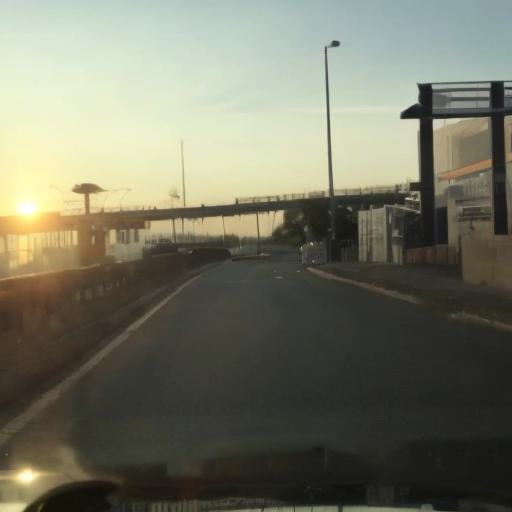}
    \includegraphics[width=0.15\textwidth]{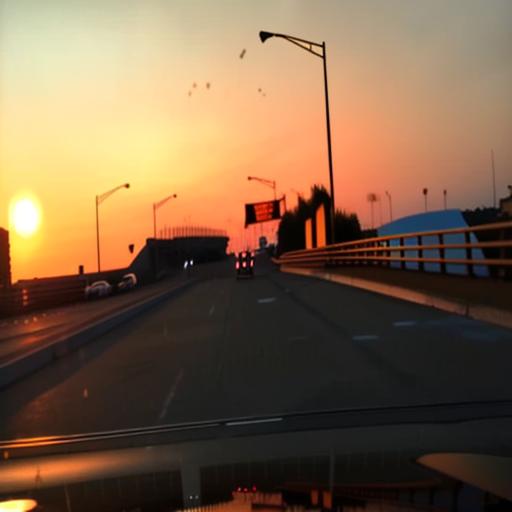}
    \\
    \raisebox{0.8\height}{\makebox[0.9\textwidth]{\scriptsize daytime→\textcolor{blue}{dawn}/ highway / clear  }}
    \\
    \raisebox{0.2\height}{\makebox[0.03\textwidth]{\rotatebox{90}{\makecell{\scriptsize 2 domain}}}}
    \includegraphics[width=0.15\textwidth]{images/domain_translation/bdd100k/2domain/source.png}
    \includegraphics[width=0.15\textwidth]{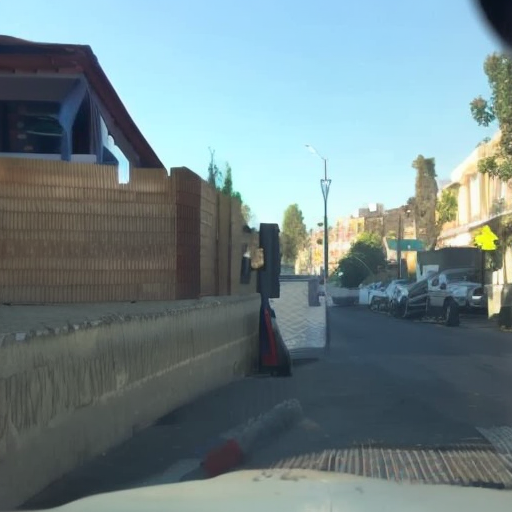}
    \includegraphics[width=0.15\textwidth]{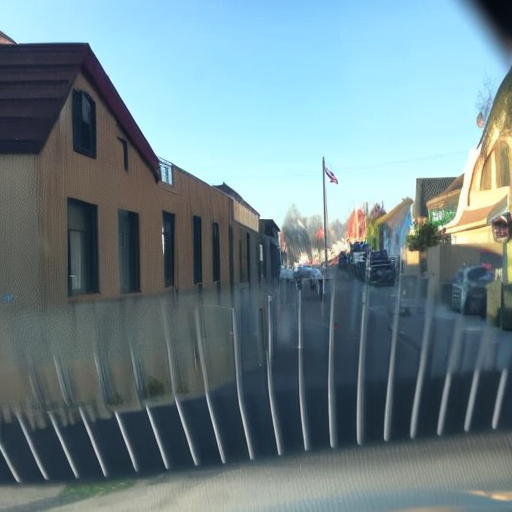}
    \includegraphics[width=0.15\textwidth]{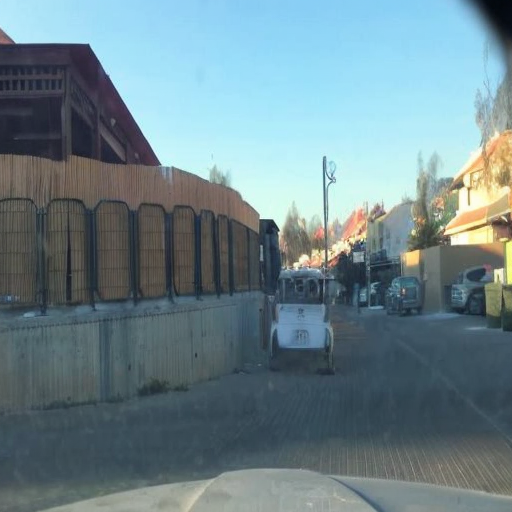}
    \includegraphics[width=0.15\textwidth]{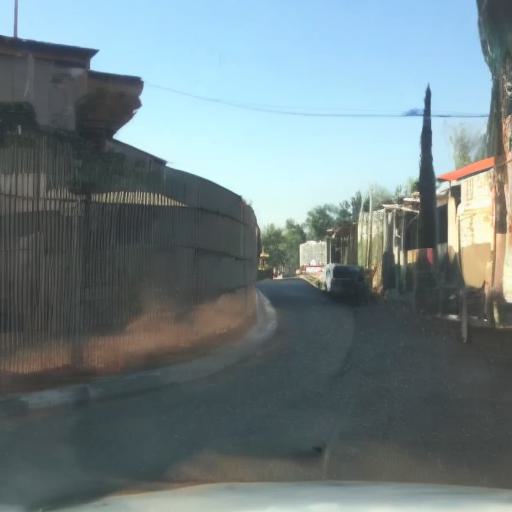}
    \includegraphics[width=0.15\textwidth]{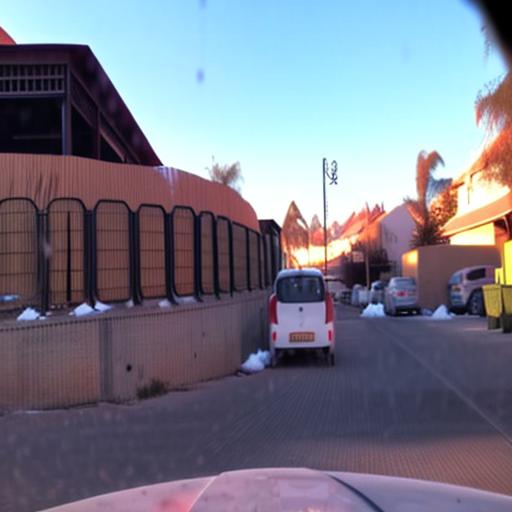}
    \\
    \raisebox{0.8\height}{\makebox[0.9\textwidth]{\scriptsize      daytime→\textcolor{blue}{dawn}/ residential / clear→\textcolor{blue}{snowy}}}
    \\
    \raisebox{0.2\height}{\makebox[0.03\textwidth]{\rotatebox{90}{\makecell{\scriptsize 3 domain}}}}
    \includegraphics[width=0.15\textwidth]{images/domain_translation/bdd100k/3domain/source.png}
    \includegraphics[width=0.15\textwidth]{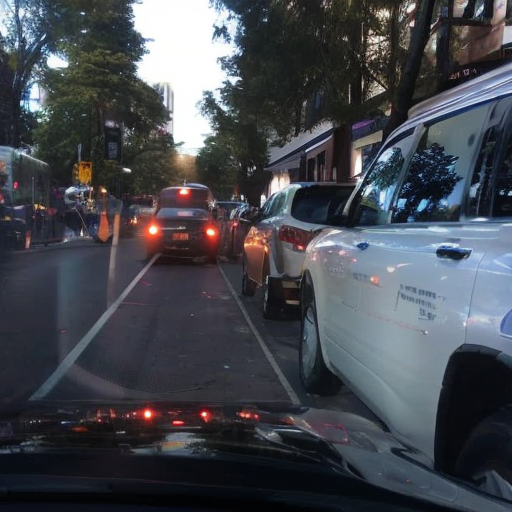}
    \includegraphics[width=0.15\textwidth]{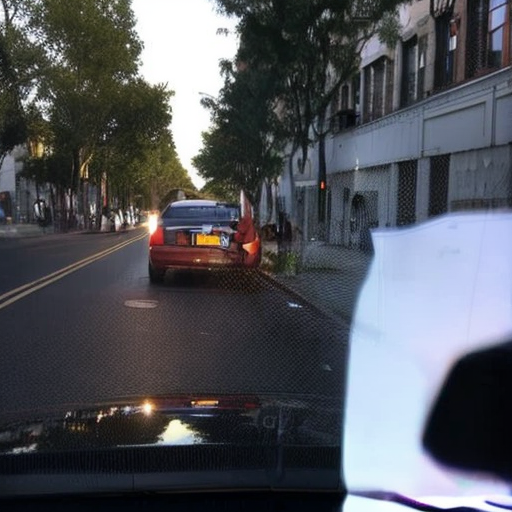}
    \includegraphics[width=0.15\textwidth]{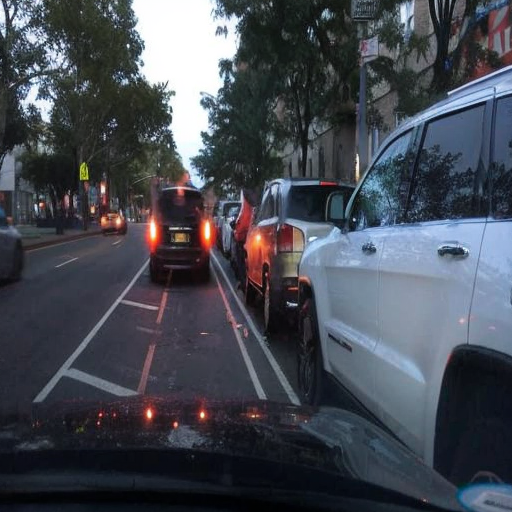}
    \includegraphics[width=0.15\textwidth]{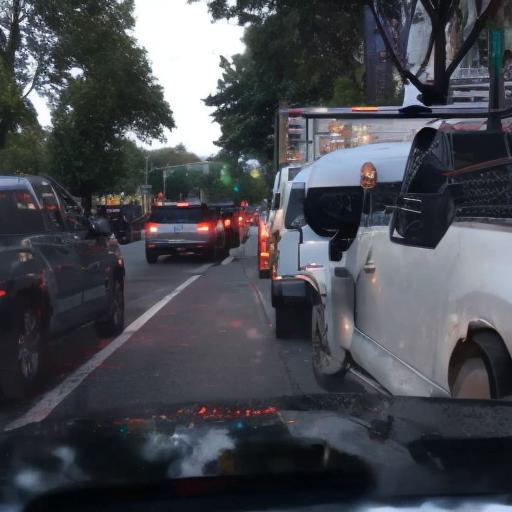}
    \includegraphics[width=0.15\textwidth]{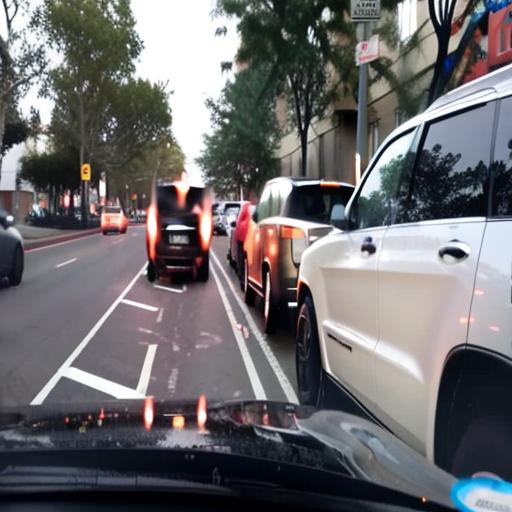}
    \\
    \raisebox{0.8\height}{\makebox[0.9\textwidth]{\scriptsize  dawn→\textcolor{blue}{daytime}  / citystreet→\textcolor{blue} {highway} /clear→\textcolor{blue}{rainy}    }}
    \\
\end{minipage}%
}
\vspace{-5pt}
    \caption{\textbf{Qualitative evaluation for multi-domain image-to-image translation methods on BDD100K.}}
    \label{fig:domain_translation_bdd100k}
\vspace{-10pt}
\end{figure*}

%% file: figure/figure_3_ablation_celeba.tex
\begin{figure}[t]
\centering
\resizebox{1\columnwidth}{!}{%
\begin{minipage}{\columnwidth}
   \centering
    \makebox[0.028\columnwidth]{}
    \makebox[0.15\columnwidth]{ \scriptsize Source}
    \makebox[0.15\columnwidth]{\scriptsize Baseline}
    \makebox[0.15\columnwidth]{\scriptsize +CLIP-Full}
    \makebox[0.15\columnwidth]{\scriptsize +GLIP-Adapter}
    \makebox[0.15\columnwidth]{\scriptsize +MCG}
    \makebox[0.15\columnwidth]{\scriptsize +DDIM(Ours)}
    \\

    \raisebox{0.3\height}{\makebox[0.03\columnwidth]{\rotatebox{90}{\makecell{\scriptsize CelebA}}}}
    \includegraphics[width=0.15\columnwidth]{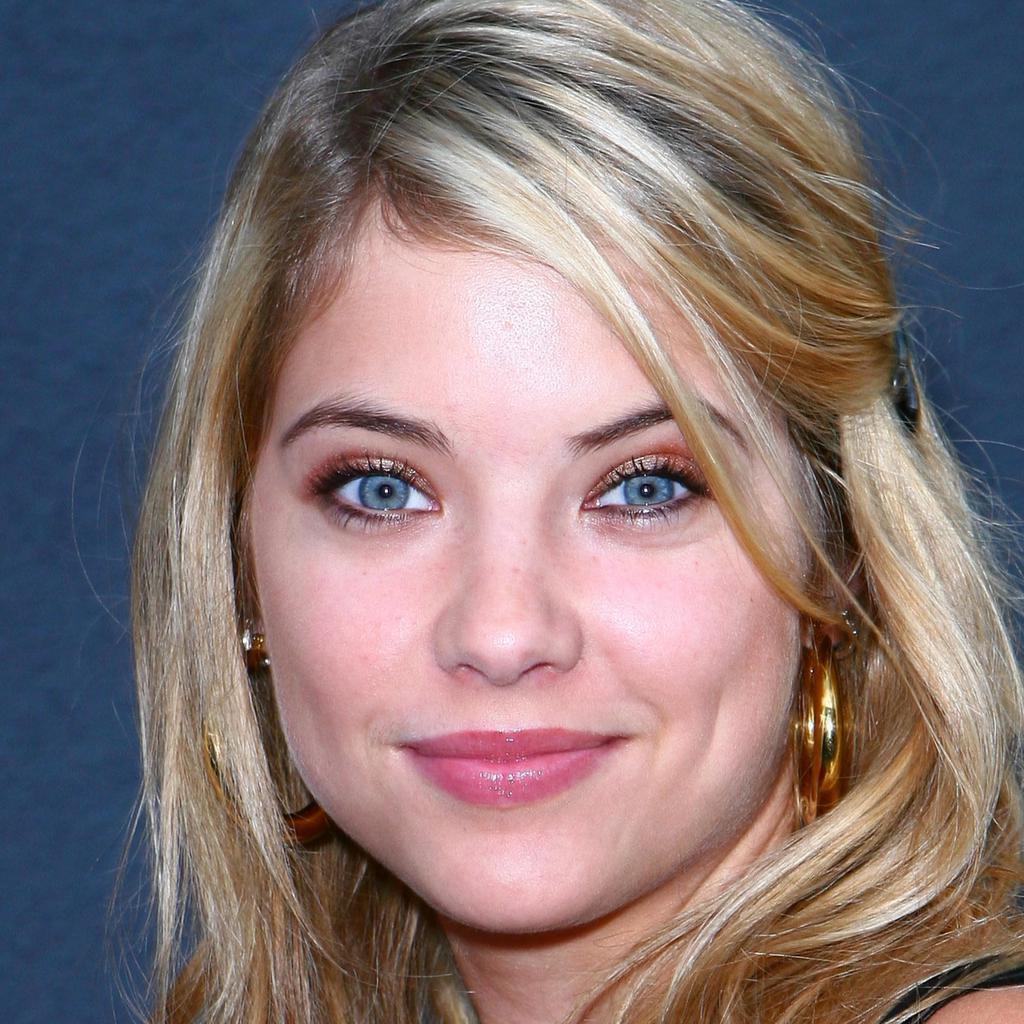}
    \includegraphics[width=0.15\columnwidth]{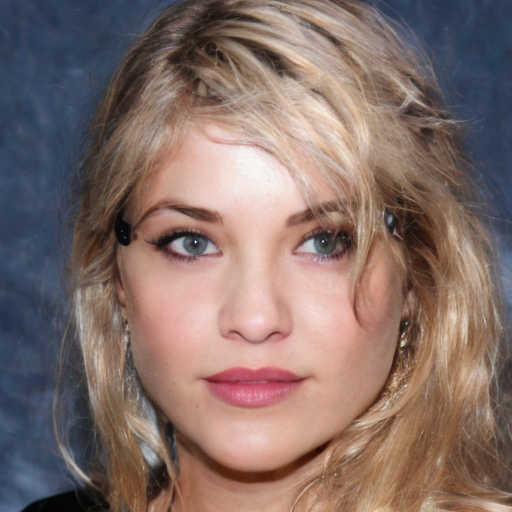}
    \includegraphics[width=0.15\columnwidth]{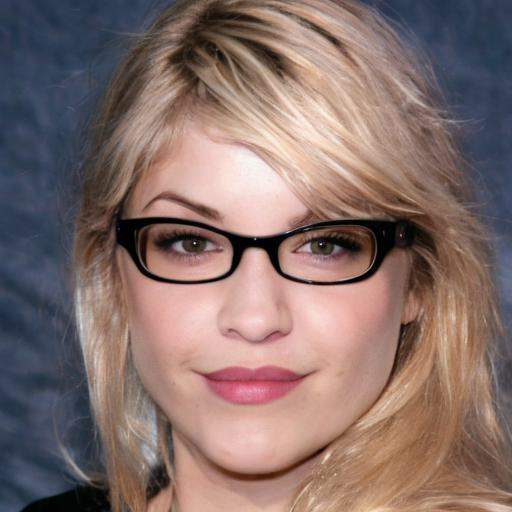}
    \includegraphics[width=0.15\columnwidth]{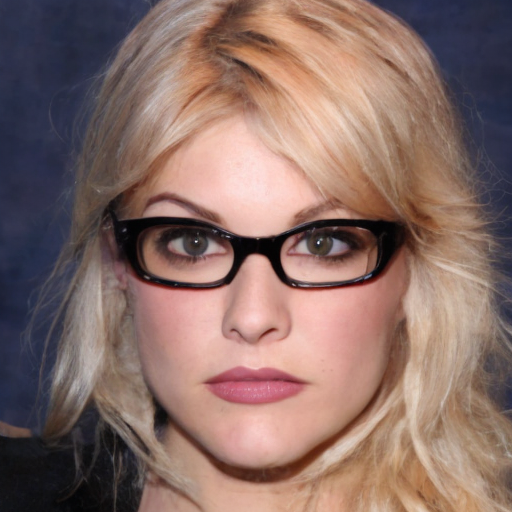}
    \includegraphics[width=0.15\columnwidth]{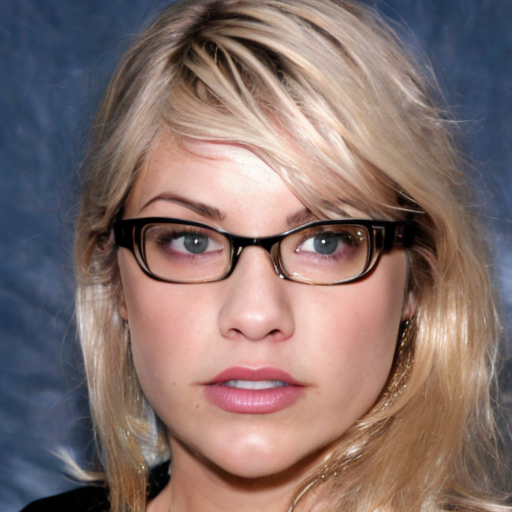}
    \includegraphics[width=0.15\columnwidth]{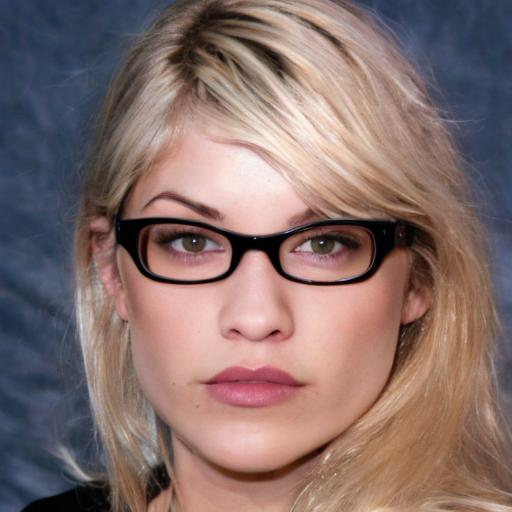}
    \\
   
    \parbox{5cm}{
    {\scriptsize female / thirties/ mild smile→\textcolor{blue}{serious face} /\\ no glasses→\textcolor{blue}{black glasses} / no bangs→\textcolor{blue}{long bangs}}
    }

     
    \raisebox{0.1\height}{\makebox[0.03\columnwidth]{\rotatebox{90}{\makecell{\scriptsize BDD100K}}}}
    \includegraphics[width=0.15\columnwidth]{images/ablation/bdd100k/3domain/source.png}
    \includegraphics[width=0.15\columnwidth]{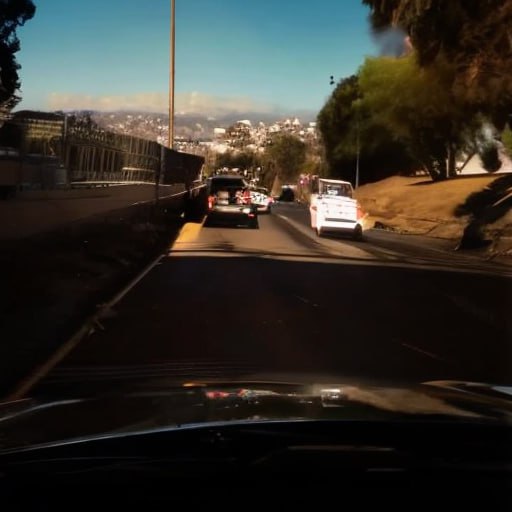}
    \includegraphics[width=0.15\columnwidth]{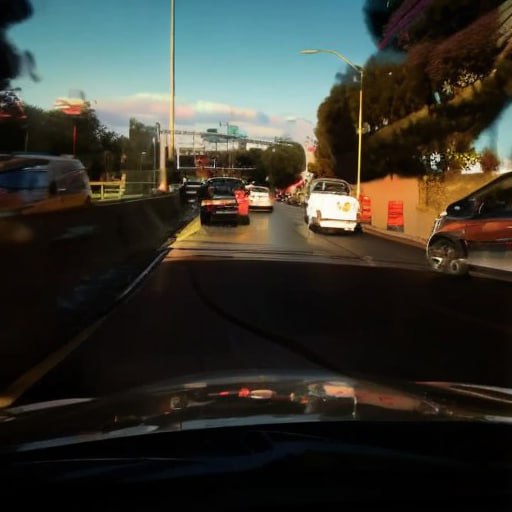}
    \includegraphics[width=0.15\columnwidth]{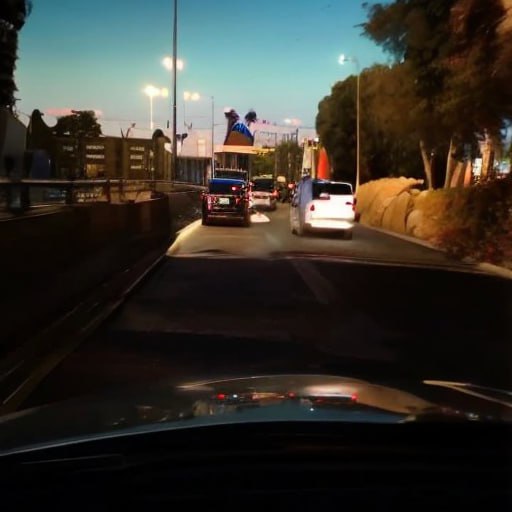}
    \includegraphics[width=0.15\columnwidth]{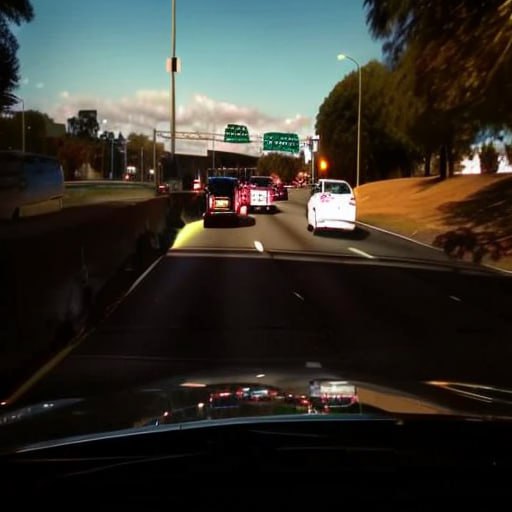}
    \includegraphics[width=0.15\columnwidth]{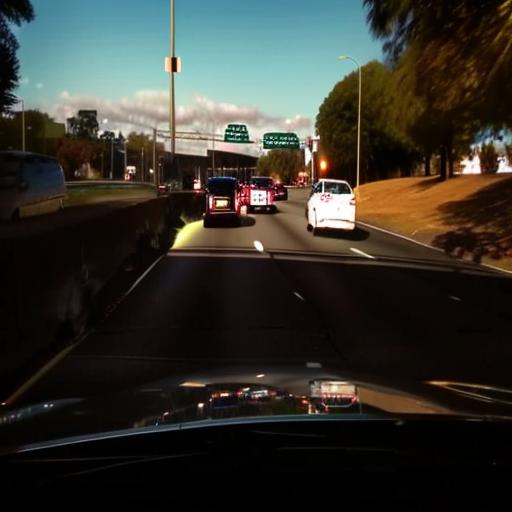}
    \\
    \raisebox{0.8\height}{\makebox[0.9\columnwidth]{\scriptsize  daytime→\textcolor{blue}{dawn} /highway→\textcolor{blue}{residential}/clear→\textcolor{blue}{overcast}  }}
    \\

    \end{minipage}
    }
    \caption{\textbf{The ablation study of multi-domain image-to-image translation methods }}
    \label{fig:ablation}
\end{figure}

%% file: table/table_experiment_3_ablation.tex
\renewcommand{\arraystretch}{1.5} 
\begin{table*}[t]\centering
    \resizebox{0.98\textwidth}{!}{
    \large
    \begin{tabular}{l|rrr|rrrr|r}
    \toprule
    \multicolumn{9}{c}{\textbf{CelebA}} \\ 
    \midrule
        &   \multicolumn{3}{c|}{} & \multicolumn{4}{c}{Background Preservation}& \\
        Method & \makecell{FID$\downarrow$} &  \makecell{FID\textsubscript{clip}$\downarrow$} &  \makecell{Structure\\Distance$\downarrow$} &  \makecell{PSNR$\uparrow$} &  \makecell{LPIPS$\downarrow$} &  \makecell{MES$\downarrow$} &  \makecell{SSIM$\uparrow$} &  \makecell{ClIP Sim$\uparrow$}  \\
        \midrule
        Baseline & 47.22 & 13.55 & 0.0789 & 15.99 & 0.39 & 0.033 & 0.56& 17.74\\ 
        +CLIP-Full $\Delta $  & -1.35 & -1.36 & -0.0032 & +0.22 & -0.0099 & -0.0072 & +0.05&+3.52 \\ 
        +DinoV2-Full $\Delta $ & -0.02 & +0.09 & -0.0002 & +0.09 & -0.0132 & +0.0021 & +0.01& +0.73\\
        +GLIP-Adapter $\Delta $ & -0.02 & -0.06 & -0.0007 &0.00  &  0.0000& -0.0017 & 0.00& +0.15\\ 
        +MCG $\Delta $ & -0.57 & -0.24 & -0.0013 & +0.18 & -0.0169 & -0.002 & +0.06&+0.91 \\ 
        +DDIM $\Delta $ & - 0.08 & - 0.58 & - 0.0002 & + 0.10 & - 0.0008 & - 0.0001 & + 0.14&+0.19 \\ \hline
        Ours & 45.18 & 11.97 & 0.0738 & 16.48 & 0.32 & 0.0318 & 0.7&23.24
        \\ 
        \bottomrule

    \toprule
    \multicolumn{9}{c}{\textbf{BDD100K}} \\ 
    \midrule
        &   \multicolumn{3}{c|}{} & \multicolumn{4}{c}{Background Preservation}& \\
        Method & \makecell{FID$\downarrow$} &  \makecell{FID\textsubscript{clip}$\downarrow$} &  \makecell{Structure\\Distance$\downarrow$} &  \makecell{PSNR$\uparrow$} &  \makecell{LPIPS$\downarrow$} &  \makecell{MES$\downarrow$} &  \makecell{SSIM$\uparrow$}&\makecell{ClIP Sim$\uparrow$}  \\
        \midrule
Baseline & 54.62 & 12.71 & 0.0652 & 17.34 & 0.32 & 0.0219 & 0.56 & 16.13\\ 
+CLIP-Full $\Delta $  & -10.14 & -3.16 & -0.0045 & +2.51 & -0.03 & +0.0010 & -0.03 & +1.37\\ 
+DinoV2-Full  $\Delta $   & -0.19  & +0.20 & -0.0022 & +0.36 & -0.04 & -0.0026 & +0.01 & +0.67\\
+GLIP-Adapter $\Delta $& -0.26  & -0.14 & -0.0021 & +0.02 & -0.02 & -0.0015 &  0.01 & +0.17\\ 
+MCG    $\Delta $    & -7.03  & -0.57 & -0.0040 & +1.88 & -0.09 & -0.0032 & +0.06 & +0.27\\ 
+DDIM  $\Delta $     & -0.98  & -0.08 & -0.0020 & +0.30 & -0.08 & -0.0031 & +0.14 & +0.37\\ \hline
Ours        & 41.07  & 7.64  & 0.0471  & 21.96 & 0.21 & 0.0096  & 0.71 & 23.24\\

        \bottomrule
        
    \end{tabular}
    }
\vspace{-5pt}
\caption{\textbf{Ablation study results showing performance metric variations in the multi-domain i2i translation methods.}
 $\Delta$ denotes the performance difference relative to the previous experiment.}
\label{tab:ablation}
\vspace{-10pt}    
\end{table*}

%% file: figure/figure_9_animal_face.tex
\begin{figure*}[t] 
\centering
    \includegraphics[width=\textwidth]{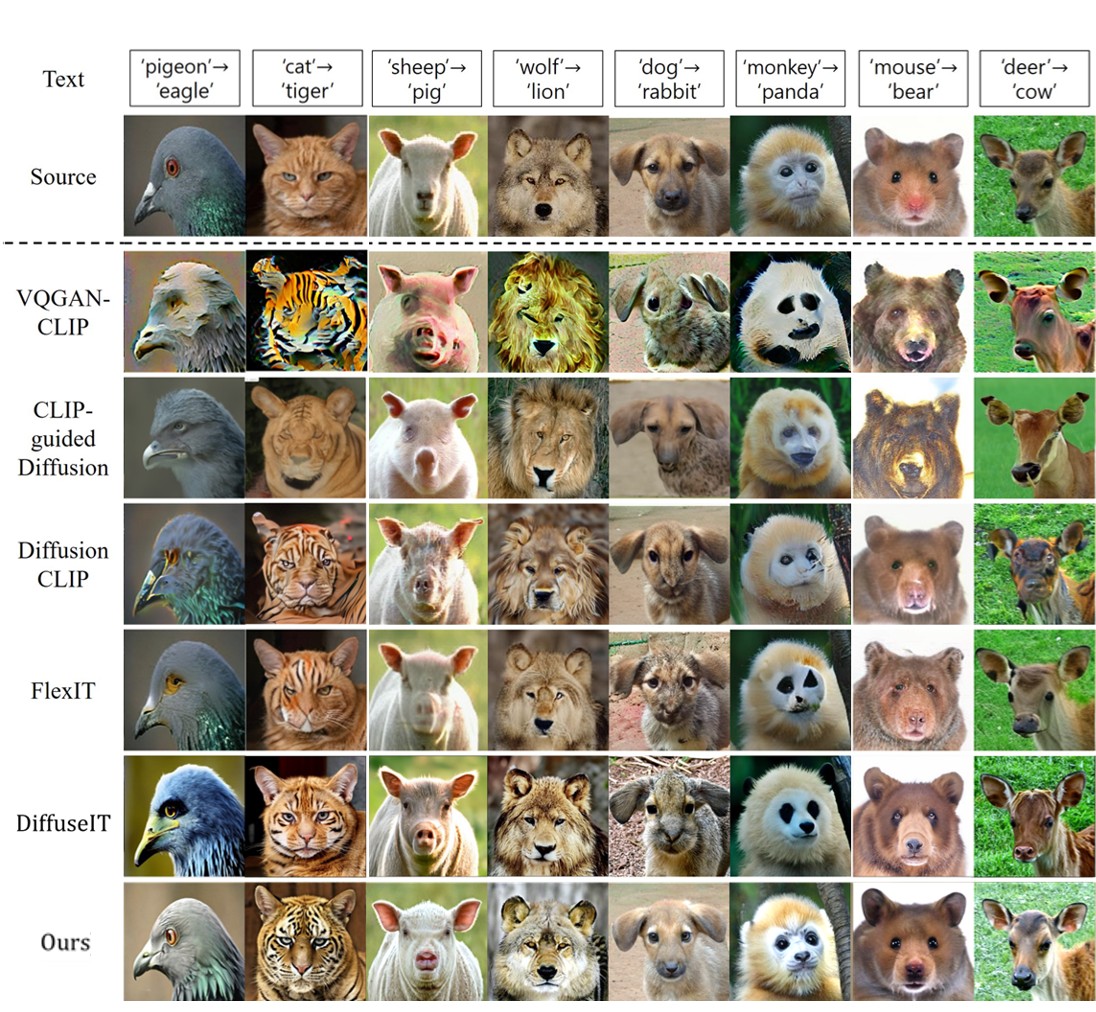}
    \caption{\textbf{Qualtitative comparison}\textbf{ with text-guided translation models} on \textbf{Animal Faces} dataset, such as DiffuseIT. Given limitations in computational resources, we leveraged the experimental setup used in DiffuseIT, training our model under identical conditions to compare results with previously evaluated models. We selected Animals Faces dataset for comparison to showcase the effectiveness of our method on more diverse datasets. Our model effectively preserves the structural content from the source image while performing accurate translations. } 
    \label{fig:animals}
\end{figure*}

%% file: figure/figure_5_scale_single_celeba.tex
\begin{figure*}[t]
\centering
\resizebox{\textwidth}{!}{%
\begin{minipage}{\textwidth}
    \centering
    \makebox[0.15\textwidth]{\scriptsize Source}
    \makebox[0.15\textwidth]{\scriptsize Target}
    \makebox[0.15\textwidth]{\scriptsize $s$ = 10}
    \makebox[0.15\textwidth]{\scriptsize $s$ = 20}
    \makebox[0.15\textwidth]{\scriptsize $s$ = 30}
    \\
    \includegraphics[width=0.15\textwidth]{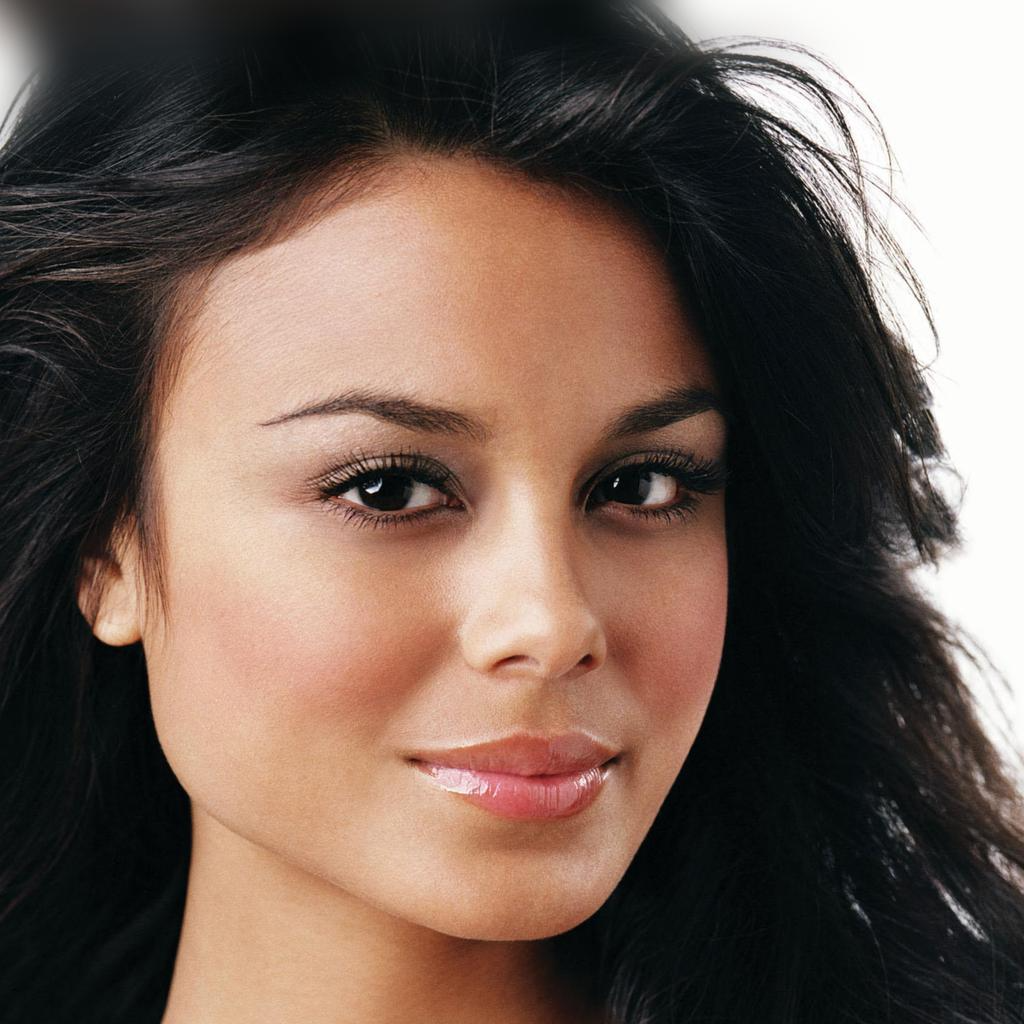}
    \includegraphics[width=0.15\textwidth]{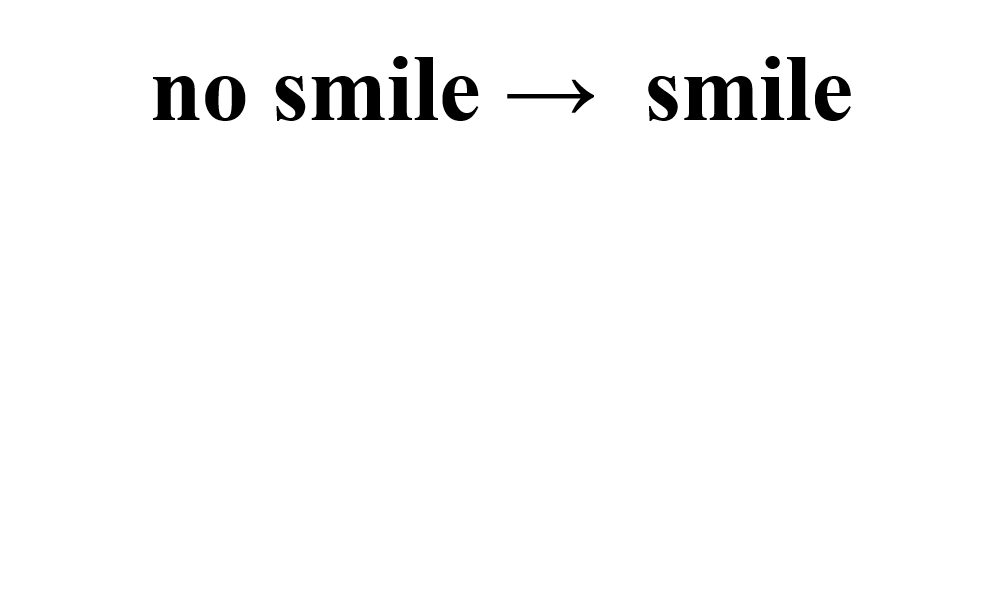}
    \includegraphics[width=0.15\textwidth]{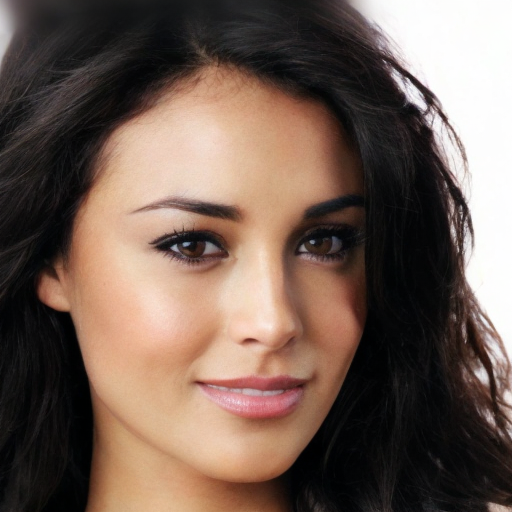}
    \includegraphics[width=0.15\textwidth]{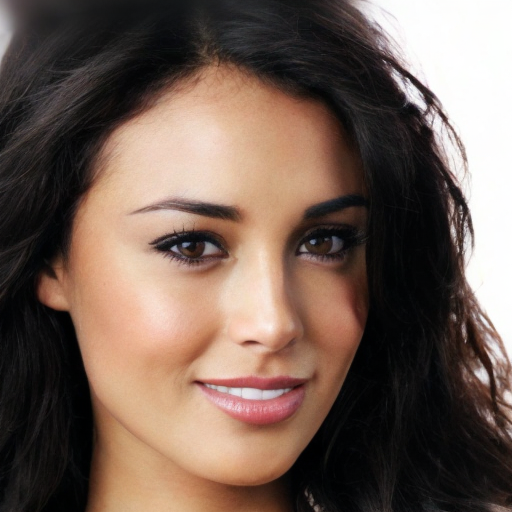}
    \includegraphics[width=0.15\textwidth]{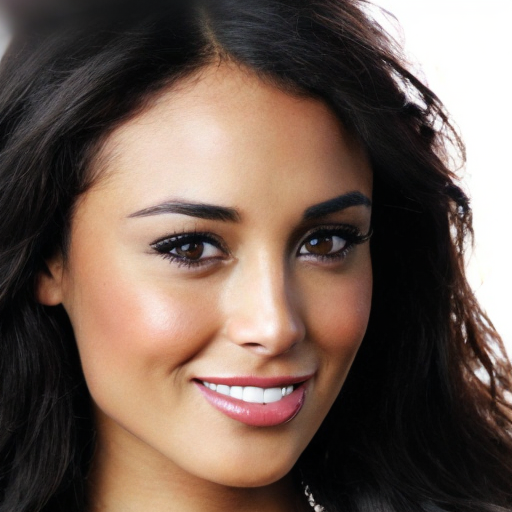}
    \\
    \includegraphics[width=0.15\textwidth]{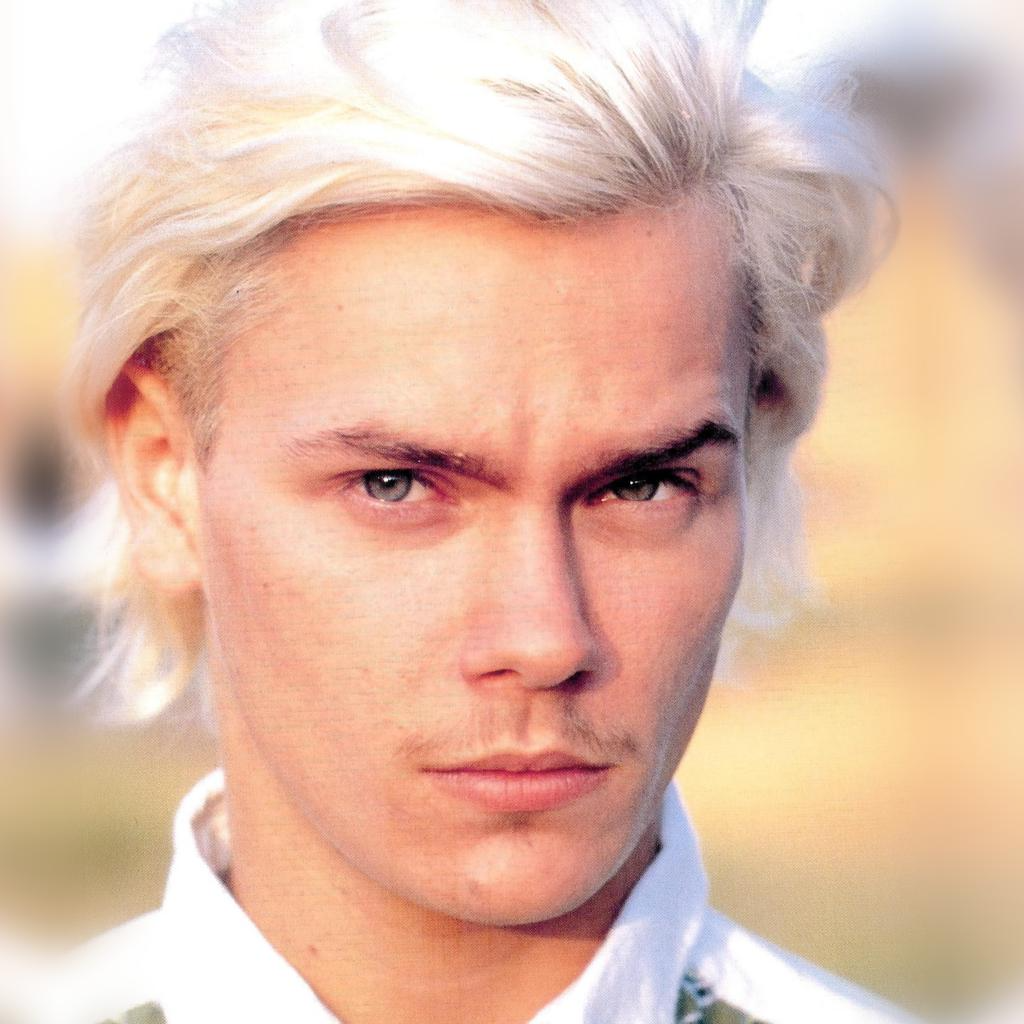}
    \includegraphics[width=0.15\textwidth]{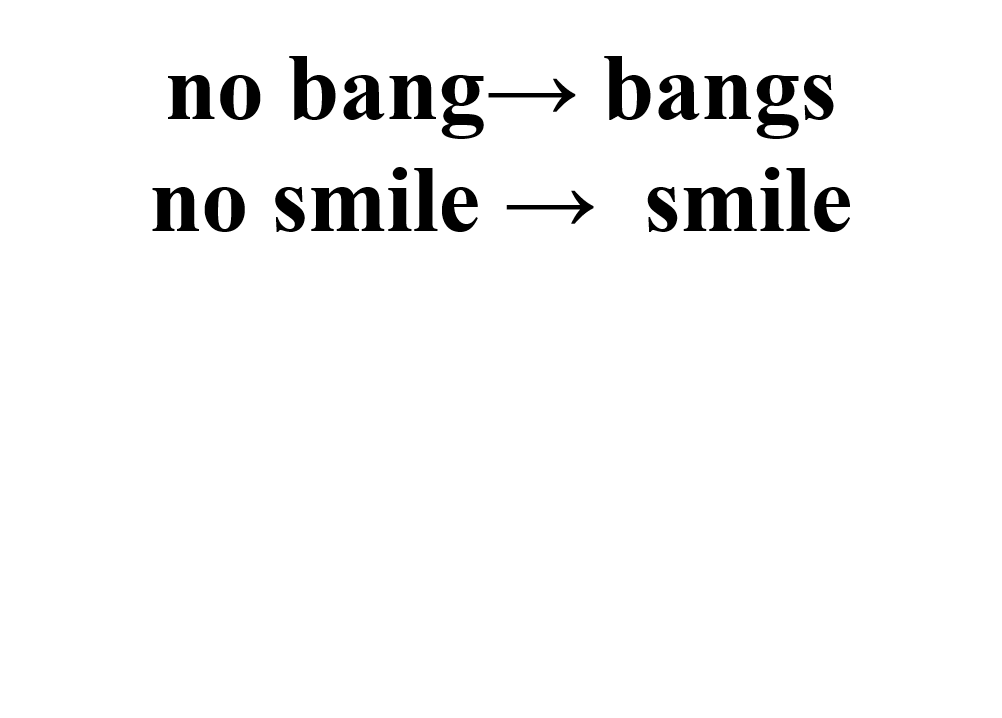}
    \includegraphics[width=0.15\textwidth]{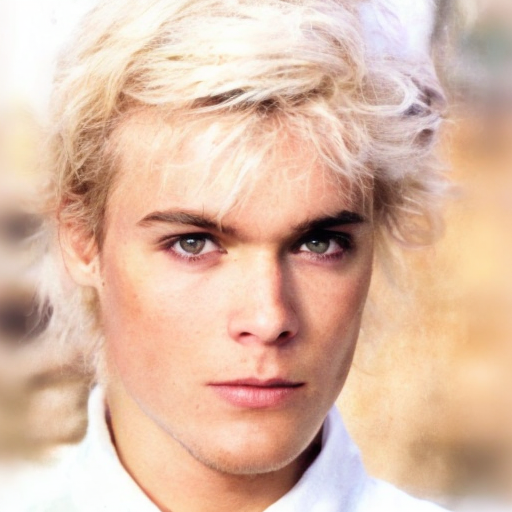}
    \includegraphics[width=0.15\textwidth]{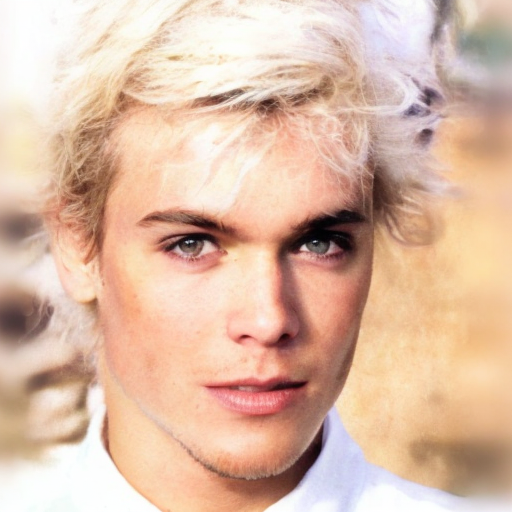}
    \includegraphics[width=0.15\textwidth]{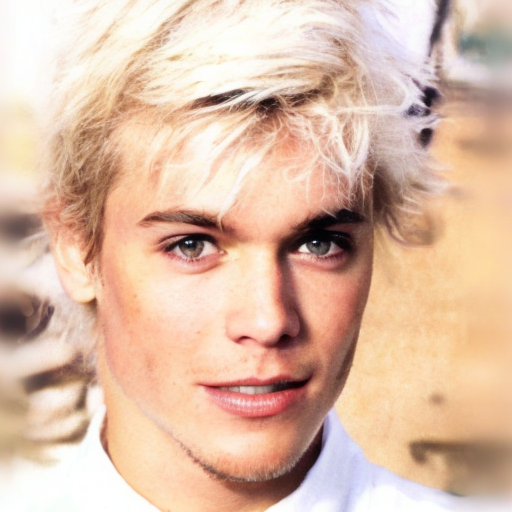}
    \\
    \includegraphics[width=0.15\textwidth]{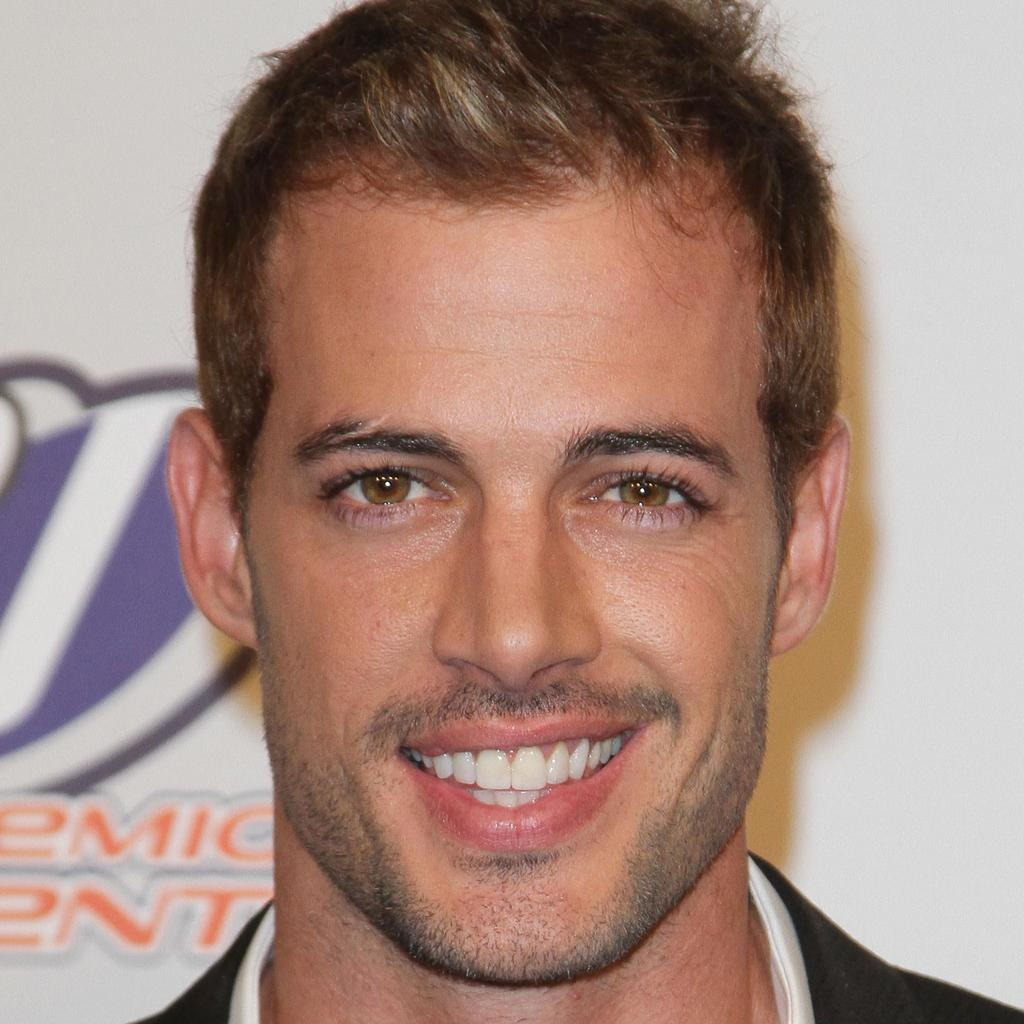}
    \includegraphics[width=0.15\textwidth]{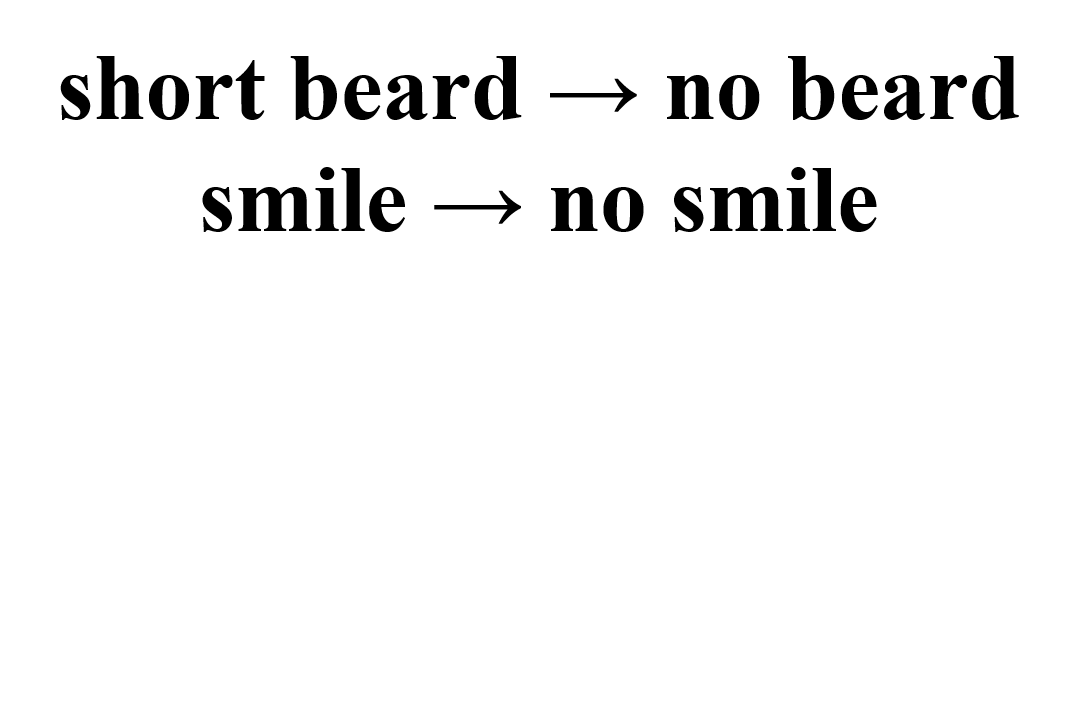}
    \includegraphics[width=0.15\textwidth]{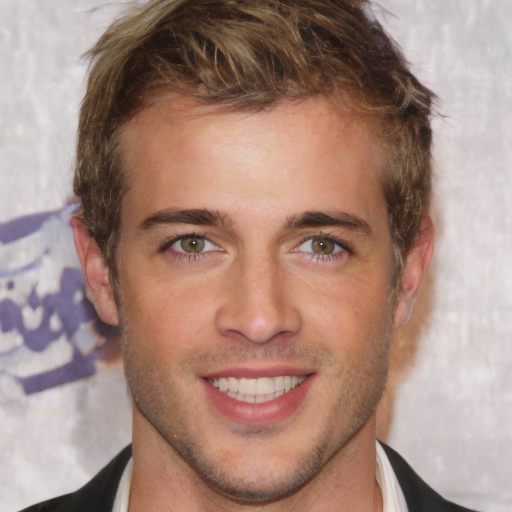}
    \includegraphics[width=0.15\textwidth]{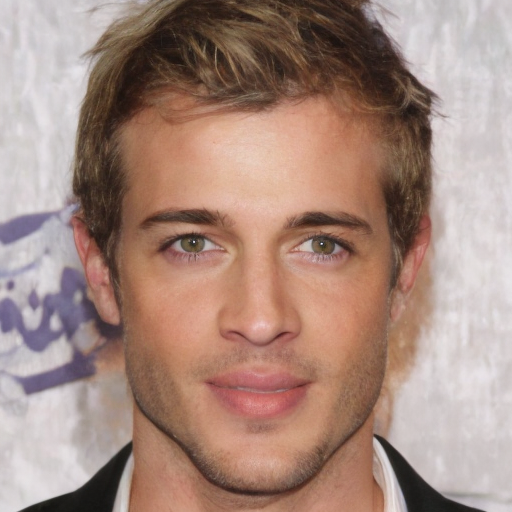}
    \includegraphics[width=0.15\textwidth]{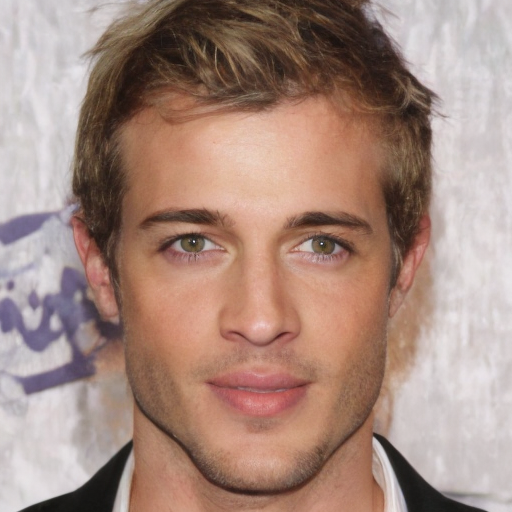}
    \\
\end{minipage}} 
    \caption{\textbf{Variations in image translation results on CelebA across the translation scale $s$.} As the translation scale increases, the visual results show stronger translations in each domains guided by the target.}
    \label{fig:scale_single_celeba}
\end{figure*}

%% file: figure/figure_8_scale_multi_bdd100k.tex
\begin{figure*}[t]
\centering
\resizebox{\textwidth}{!}{%
\begin{minipage}{\textwidth}
    \centering
    \makebox[0.15\textwidth]{\scriptsize Source}
    \makebox[0.15\textwidth]{\scriptsize Target}
    \makebox[0.15\textwidth]{\scriptsize $s_1$ = 10 , $s_2$ = 10}
    \makebox[0.15\textwidth]{\scriptsize $s_1$ = 30 , $s_2$ = 10}
    \makebox[0.15\textwidth]{\scriptsize $s_1$ = 10 , $s_2$ = 30}
    \\
    \includegraphics[width=0.15\textwidth]{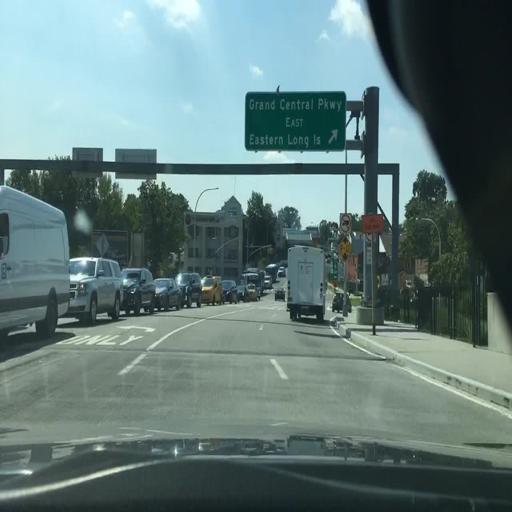}
    \includegraphics[width=0.15\textwidth]{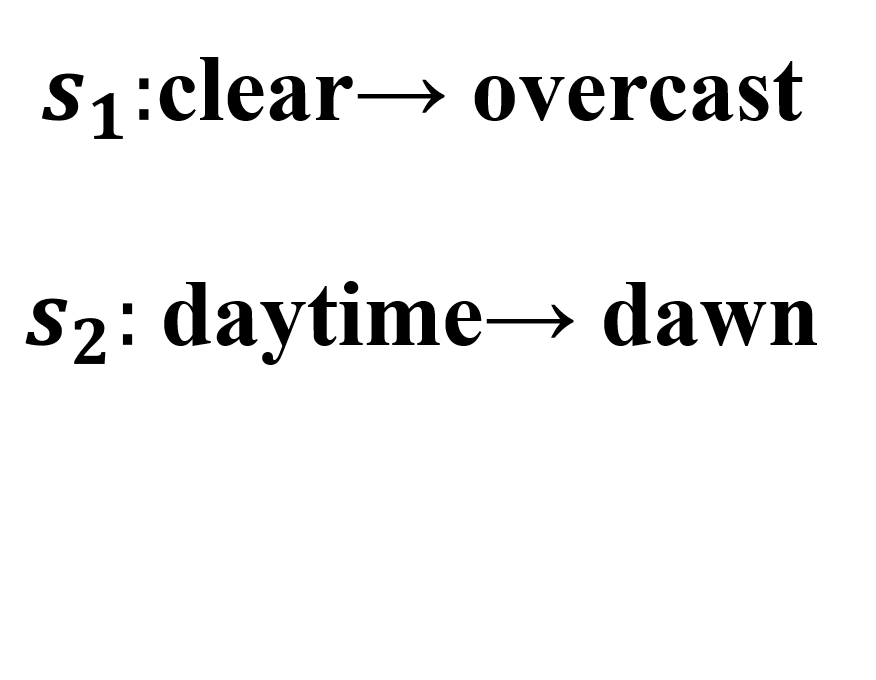}
    \includegraphics[width=0.15\textwidth]{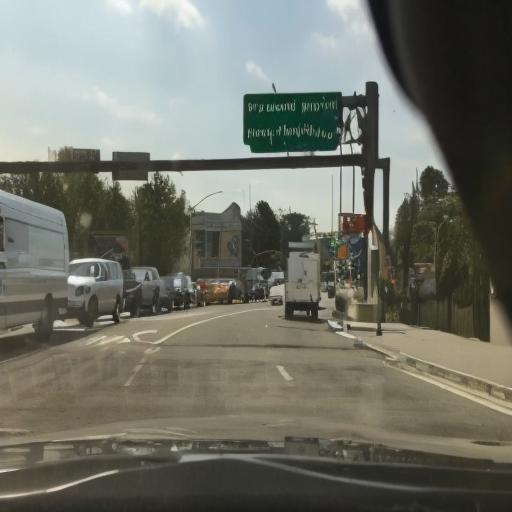}
    \includegraphics[width=0.15\textwidth]{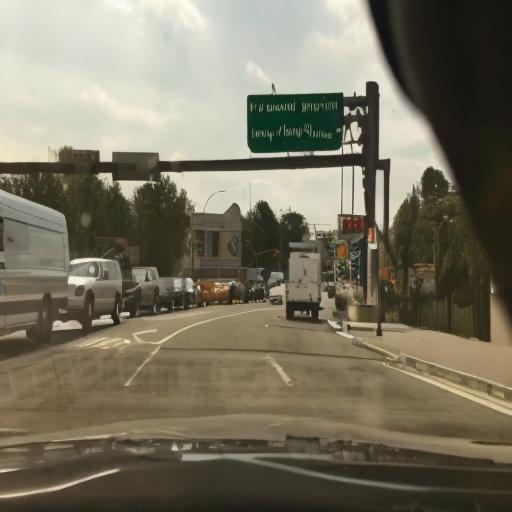}
    \includegraphics[width=0.15\textwidth]{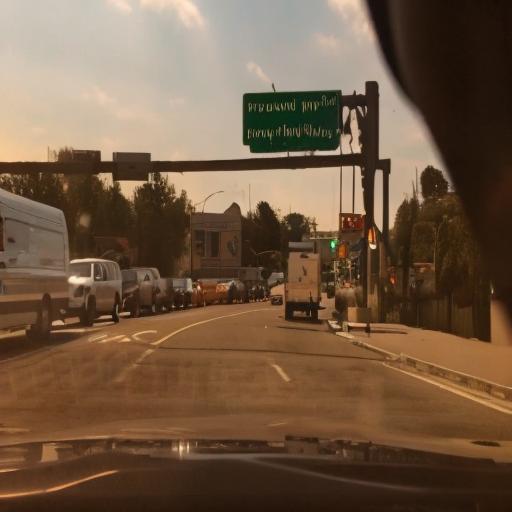}
    \\
    \includegraphics[width=0.15\textwidth]{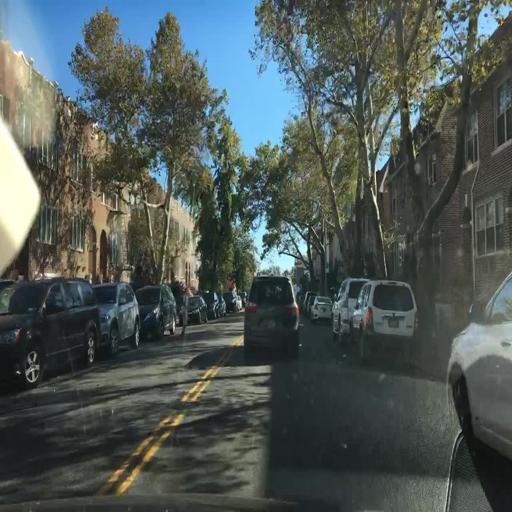}
    \includegraphics[width=0.15\textwidth]{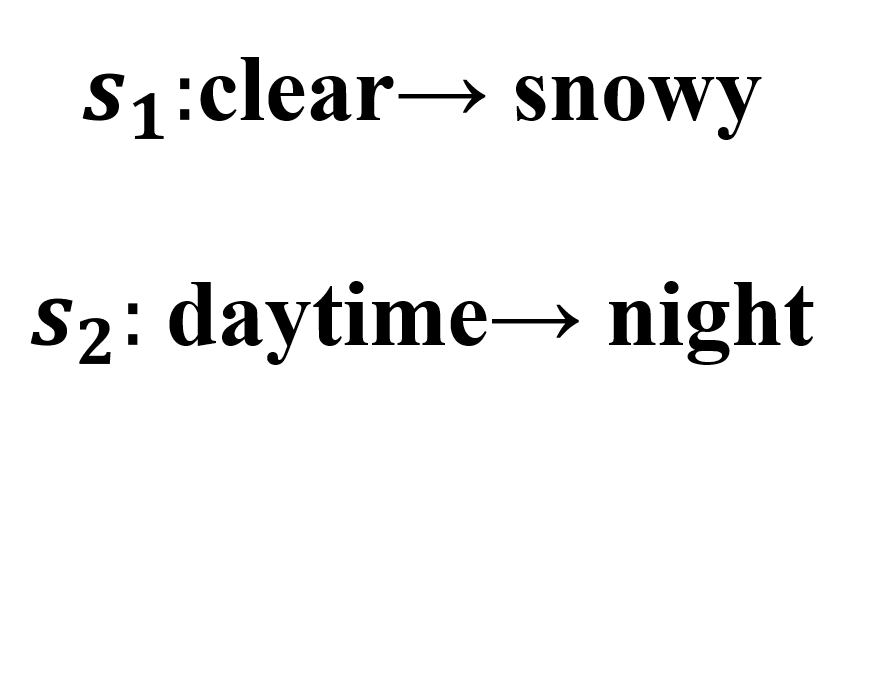}
    \includegraphics[width=0.15\textwidth]{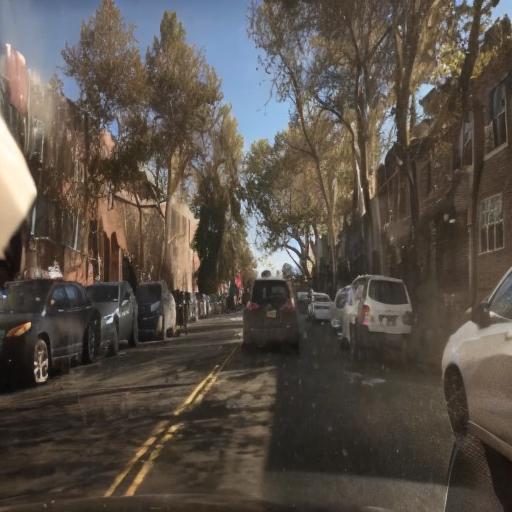}
    \includegraphics[width=0.15\textwidth]{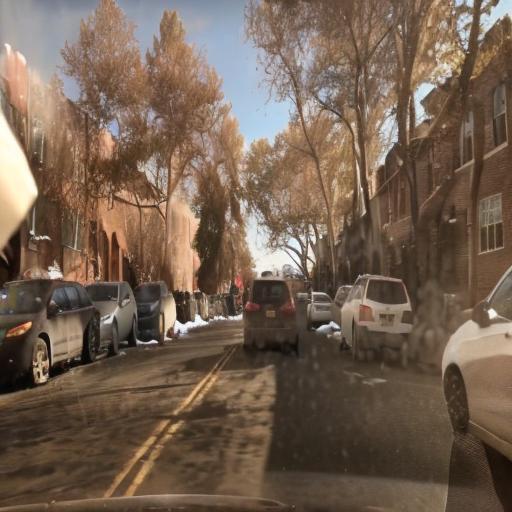}
    \includegraphics[width=0.15\textwidth]{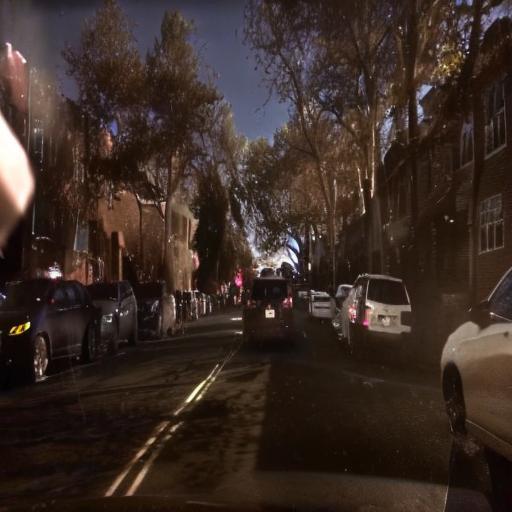}
    \\
    \includegraphics[width=0.15\textwidth]{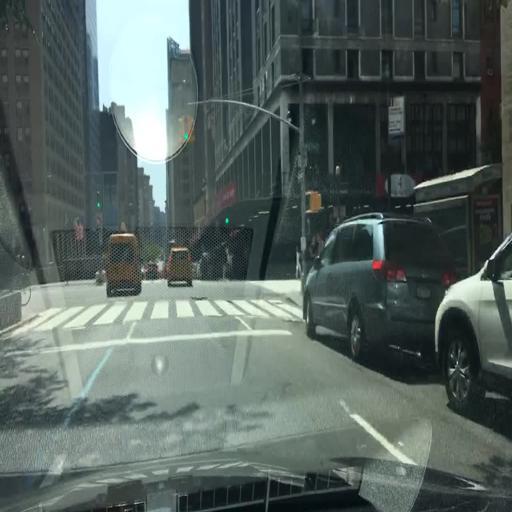}
    \includegraphics[width=0.15\textwidth]{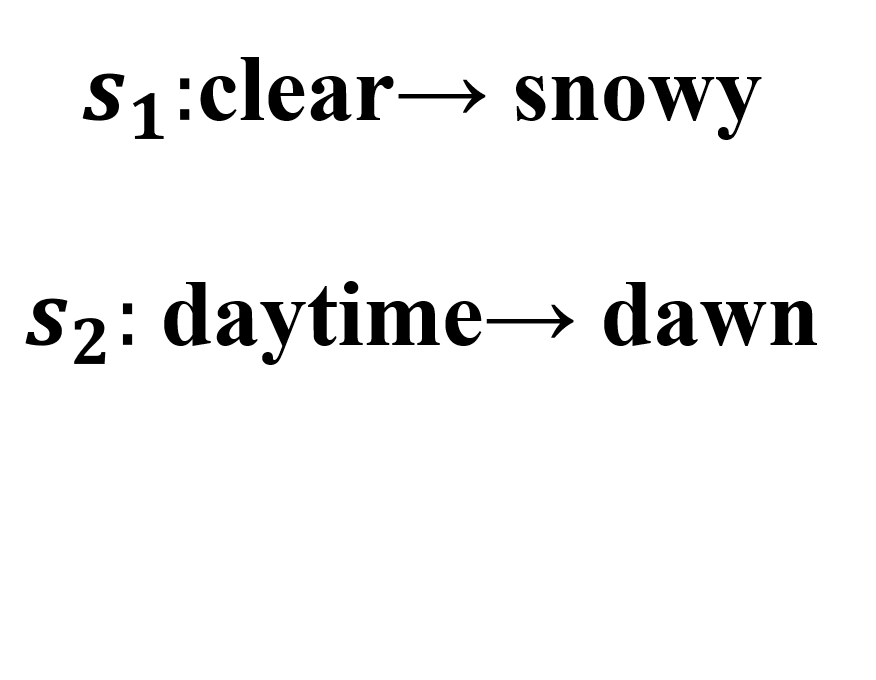}
    \includegraphics[width=0.15\textwidth]{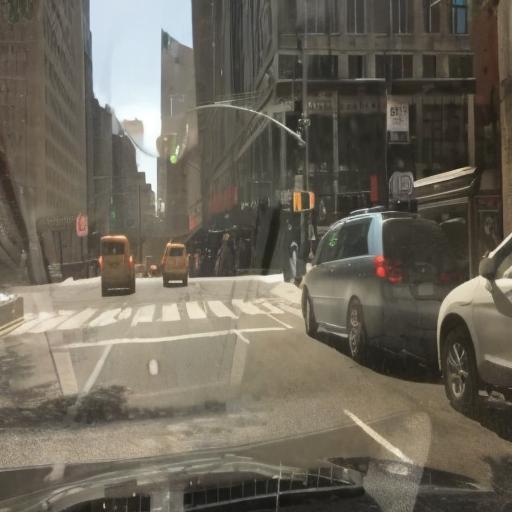}
    \includegraphics[width=0.15\textwidth]{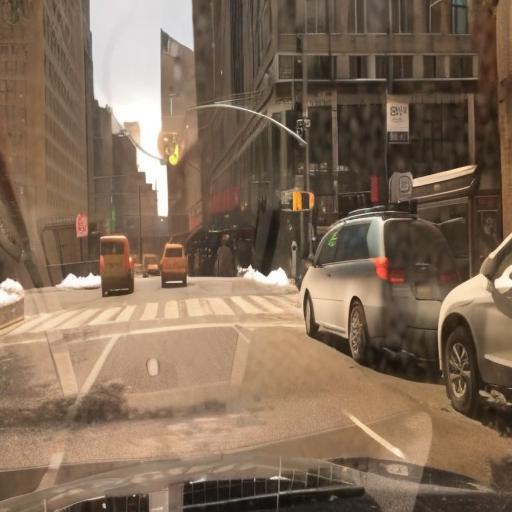}
    \includegraphics[width=0.15\textwidth]{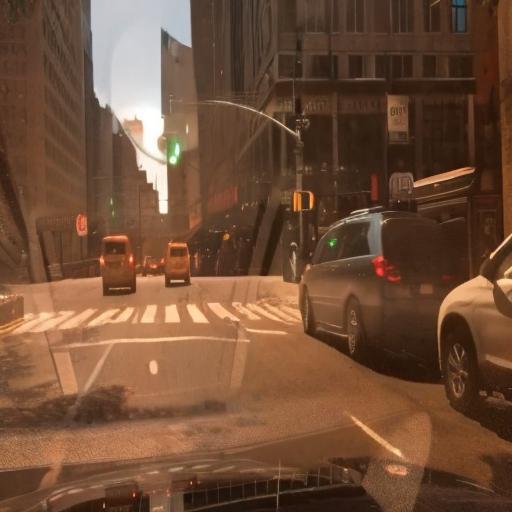}
    \\
\end{minipage}
}    
    \caption{\textbf{Variations in image translation results on BDD100K across the differential translation scales $s_1$, $s_2$.} The two differential translation scales, $s_1$ and $s_2$, control the degree of translation for the weather and time of day domains, respectively.}
    \label{fig:scale_multi_bdd100k}
\end{figure*}